\documentclass[journal,twocolumn]{IEEEtran}

\usepackage{graphicx}
\usepackage[natbib=true,style=numeric,sorting=none]{biblatex}
\usepackage{cancel}
\usepackage{float}

\usepackage{multirow}
\usepackage{array, makecell, rotating}
\usepackage{tabularx}
\newcommand\nocell[1]{\multicolumn{#1}{c|}{}}

\begin{document}

%
\title{What Image Features Boost Housing Market Predictions?}
%
%
%

\author{Zona Kostic,~\IEEEmembership{Harvard University,}
        Aleksandar Jevremovic,~\IEEEmembership{Member,~IEEE}
\thanks{Z. Kostic is with the Harvard John A. Paulson School Of Engineering and Applied Sciences, Cambridge, MA, USA. A. Jevremovic is with the Faculty of Informatics and Computing,  Singidunum University, Belgrade, Serbia}
\thanks{Manuscript received ---}}


\markboth{IEEE TRANS MULTIMEDIA,~Vol. 22, No. 7, July 2020, DOI: 10.1109/TMM.2020.2966890~}%
{Shell \MakeLowercase{\textit{et al.}}: Bare Demo of IEEEtran.cls for IEEE Journals}

\maketitle

\begin{abstract}

The attractiveness of a property is one of the most interesting, yet challenging, categories to model. Image characteristics are used to describe certain attributes, and to examine the influence of visual factors on the price or timeframe of the listing. In this paper, we propose a set of techniques for the extraction of visual features for efficient numerical inclusion in modern-day predictive algorithms. We discuss techniques such as Shannon's entropy, calculating the center of gravity, employing image segmentation, and using Convolutional Neural Networks. After comparing these techniques as applied to a set of property-related images (indoor, outdoor, and satellite), we conclude the following: (i) the entropy is the most efficient single-digit visual measure for housing price prediction; (ii) image segmentation is the most important visual feature for the prediction of housing lifespan; and (iii) deep image features can be used to quantify interior characteristics and contribute to captivation modeling. The set of 40 image features selected here carries a significant amount of predictive power and outperforms some of the strongest metadata predictors. Without any need to replace a human expert in a real-estate appraisal process, we conclude that the techniques presented in this paper can efficiently describe visible characteristics, thus introducing perceived attractiveness as a quantitative measure into the predictive modeling of housing.

\end{abstract}


\IEEEpeerreviewmaketitle

\section{Introduction}
\IEEEPARstart{A}{ccurate} price predictions can be used to benefit a wide range of different interest groups, including urban planners, real estate developers and investors. However, in many locations, democratizing information about market conditions is the exclusive domain of larger brokerage houses. A more accurate prediction tool could counter these monopolistic tendencies and could also benefit the broader public. With this aim, several real estate companies now provide property estimates, investing significant financial resources into improving proprietary predictive models \cite{ZillowPrizeWebpage}. Ensuring that people have a trusted way to monitor active market assets is paramount. However, market changes are not always easily predictable, and cannot be represented by a single formula. For example, in the city of Boston, MA, USA, massive foreign interest drives up prices, consequently decreasing the average lifespan of units \cite{ReportMLSPIN}. Driven by massive demand, multiple offers per listing have now become an ongoing trend, forcing immediate sales above the asking price.


Our curiosity was triggered by the {\textit{reasons}} behind this behavior. What prompts buyers to pay more for one unit compared to another of the same type, in the same location? Although a set of specific needs, such as amenities, location, budget, or time, drives every purchase, is there a common set of interior or exterior aspects that encourages buyers to put more money down and allows the property to sell faster? 

Despite the well-known importance of visual characteristics in making purchasing decisions, techniques such as Machine Learning (ML) are not able to model personal preferences efficiently \cite{1423983}. However, recent advances in computer vision modeling include smart content interpretations \cite{7937942}, thus opening a new set of transformations to account for personal tastes. Could content abstractions obtained from deep image analysis be successfully employed in (un)supervised scenarios, representing this “knowledge” in analytical formulas, and help to improve the overall predictive power of a model? 

In this paper, we investigate visual features as a factor in an improved understanding of real estate {\textit{attractiveness}}. We work with two response variables: {\textit{price}} and {\textit{dom}} (Days On the Market) and analyze approximately 20,000 properties in Massachusetts and New York, USA. The main idea is to incorporate a set of visual features into a standard ML model to account for subjectivity (what makes a good listing), competence (what makes a good purchase), or a personal taste (what buyers like). We propose a carefully chosen collection of supervised and unsupervised extraction methods, and use them to derive a set of {\textit{image features}}. 

This research was based on all types of property-related images, i.e. {\textit{indoor}}, {\textit{outdoor}}, and {\textit{satellite}}, thus including different visual perspectives enabling a unique overlook. Visual aspects such as indoor style or layout proved to be key features of the property's market value. Widely spaced, green neighborhoods and outdoor amenities speed up the purchase, whereas less attractive houses tend to be adjacent to each other, slowing down the selling process. In addition to the interior and exterior information, satellite images provide contextual clues about broader surroundings, allowing location-aware specifics to be known.

Three different techniques were employed to process the various image sources: {\textit{Shannon's entropy}} and {\textit{center of gravity}}  to quantify the information potential of an image; {\textit{image segmentation}} to extract the quality of the environment; and a {\textit{Convolutional Neural Network}} (CNN) to estimate and quantify the indoor style, categories, and amenities. Combined with basic predictors, the use of image features can augment the performance of the model, outperforming some of the strongest metadata variables. A group of 40 features was selected as the main contributors, achieving successful content interpretation with the full set of techniques proposed.

The rest of the work is organized as follows: we present related work in Section 2; image extraction techniques are described in Section 3; Section 4 describes the experiments conducted to estimate {\textit{price}} and {\textit{dom}}, the metrics used and the results obtained; we discuss the results and conclude the paper with suggestions for future work in Section 5.

\section{Related Work}
The earliest housing market models analyzed trends using regression lines \cite{Wardrip11publictransit}. Methods such as this assume that the price is a weighted sum of property characteristics, and are unable to address non-linearity or detect outliers \cite{DBLP:journals/corr/PoursaeedMB17}. Time-series models like autoregression \cite{Dubin1999} can reflect trends more adequately, and are often used to model supply and demand within a market \cite{Wang_2018}. Academic approaches to predicting home prices have traditionally relied on the hedonic approach, in which the sale price of a property is understood as a function of a finite set of characteristics \cite{Reichert2002}. However, hedonic models are restrictive, imposing uniformity of coefficients across both space and time \cite{doi:10.1177/03058298780070030601}, and time-series methods prove to be efficient only when inaccessible, proprietary data from eminent brokerage houses are employed.

ML has been adopted as a modern-day extension for predictive analytics and time-series modeling. However, Neural Networks (NN) are considered weak forecasting machines, due to their slow convergence rates and overfitting of training data \cite{Wang_2018}. If a model fails to perform accurately, different subsets of predictors can be combined to 'boost' the overall predictive power. Boosting algorithms \cite{10.1007/978-3-642-10677-4_58} have been demonstrated with the use of  combinatorial techniques for identifying important predictors in high-dimensional space, combining variables with nonlinear effects \cite{StackingEnsemble}. Consequently, boosting has great potential to give results that are better than those from an NN or other time-series forecasting model alone \cite{Yoonseok2015}.

The expansion of image-based extraction methods has reinforced the use of CNNs, and image abstractions have been successfully combined with other time-series models \cite{DBLP:journals/corr/GuerinGTN17aa}. The research community has recently shown interest in real estate appraisal problems, such as examining the influence of visual factors \cite{7937942} or deriving an unequivocal visual style for the property \cite{DBLP:journals/corr/PoursaeedMB17}. The academic community has started utilizing pictures as one of the most important factors in the real-estate valuation process, including satellite images \cite{7926625}, property-related indoor images \cite{PredictionLiYu}, and street/outdoor images \cite{2018arXiv180707155L}.

CNN models use selective attention techniques in order to understand human perception. A visual representation of what appears to be important in the image is usually presented using saliency maps \cite{zhou2015cnnlocalization}, which can be thought of as the average self-information across the content. An entropy metric as a “saliency-driven” method was successfully employed to extract informative image regions \cite{SaliencyAttention}. The features extracted from the pictures draw high-level human perceptions on well-being \cite{6875954,6909869}, and are correlated with housing market movements (e.g., estimating the effect of visible "greenness" on housing prices \cite{ijgi7030104}). Used in this way, a picture gives rich visual information about the immediate environment or larger neighborhood, utilizing various types of available image data.

In \cite{7926625}, satellite images were analyzed in the context of object detection, image segmentation, or image tracking. However, the entropy metric proposed in this paper was used to measure the level of urbanization, focusing on the immediate position of the property and the surrounding area. The entropy metric is a computationally less expensive process that provides solid and reliable quantified information taken from unstructured data.

In this paper, we combine different image processing techniques to efficiently extract the features used for modeling the attractiveness of a property. Similarly to the majority of research projects described in this section, we find that the importance of the location is crucial in predicting real-estate trends. In addition to \cite{7937942} which derives location information by employing random walks to generate housing sequences, other models use location information extracted from metadata (meaning all property-related information downloaded directly from a real-estate database). Although there are frameworks that have suggested a solely image-based approach (e.g. \cite{RichardRYAIBlueBook}), none of these has ever been applied in a house pricing scenario without fostering the importance of locality.

The next section provides a detailed description of the data, feature extraction methods, and techniques applied.


\begingroup
\setlength{\tabcolsep}{2pt} 
\renewcommand{\arraystretch}{1.1} 
\begin{table}
    \scriptsize
    \begin{tabular}{ |c|c|c| }
        \cline{2-3}
        \nocell{1} & \textbf{\# listings} & \textbf{MLS feature: feature description} \\
        \hline
        \textbf{MLS Data} & 19,942 & \multirow{12}{*}{\makecell{
        MLSNUM: reference number\\
        SOLDPRICE: final property price\\
        DOM: days on the market\\
        DTO: days to offer\\
        ADDRESS, CITY, ZIP: location features\\
        BEDS,BATHS: number of beds/baths\\
        LOTSIZE: lot size of the house\\
        GARAGE: binary garage variable\\
        AGE: how old the property is\\
        HIGHSCHOOL: school description\\
        REMARKS: property description\\
        PHOTOURL: the URLs for images
        }} \\
        \cline{1-2}
        \textbf{Train dataset} & 15,542 & \\
        \cline{1-2}
        \textbf{Test dataset} & 4,000 &  \\
        \cline{1-2}
        \textbf{Metadata (features)} & 873 & \\
        \cline{1-2}
        & & \\
        & & \\
        & & \\
        & & \\
        & & \\
        & & \\
        & & \\
        & & \\
        & & \\
        \hline
        \multicolumn{3}{|c|}{\includegraphics[width=8cm]{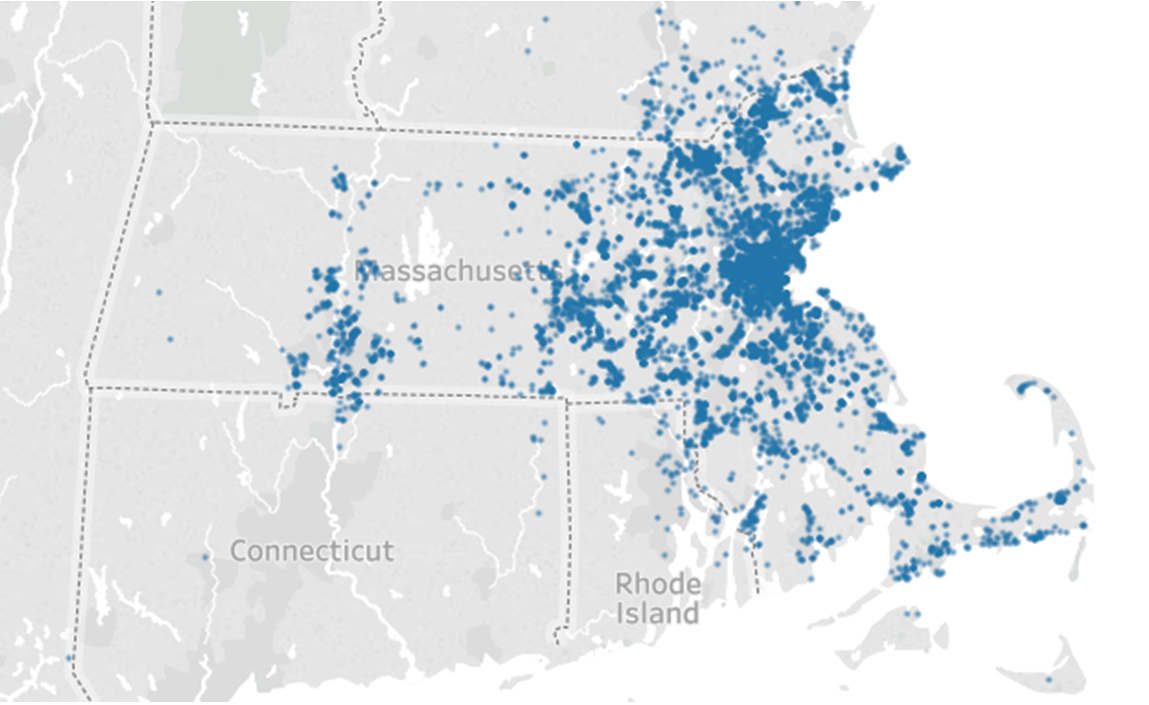}}\\
        \hline
    \end{tabular}
    \\
    \caption{MLS data, features, and location}
    \label{tab:test}
\end{table} 
\endgroup

\section{Data Analysis and Feature Extraction}
\subsection{MLS metadata}
A Multiple Listing Service (MLS) is a service allowing realtors and other realty professionals to find homes for sale, new homes, or resale homes \cite{MLSSite}. We used the MLS to obtain property-related information for almost 20,000 units in New York and Massachusetts (sold in 2016). Each property has 873 metadata descriptors. Some high-level statistics on the dataset and the location-based distribution of the houses are listed in Table 1.

Before extracting image features, we run a boosting model using all MLS predictors for both response variables: {\textit{price}} and {\textit{dom}}. Given a large set of predictors, we employ a gradient boosting algorithm and plot an importance graph to narrow down the feature space. Plotting feature importance is a conventional measure of interpretability that underlines the combination of useful predictors \cite{ViktoriyaISVFIUBF}. Figure 1 presents the importance plot with the top 40 numerical features of the MLS generated for both response variables. The figure provides a high level of insight into the behavior of the model, taking into an account all interactions with other features.

\begin{figure*}
    \includegraphics[width=\textwidth]{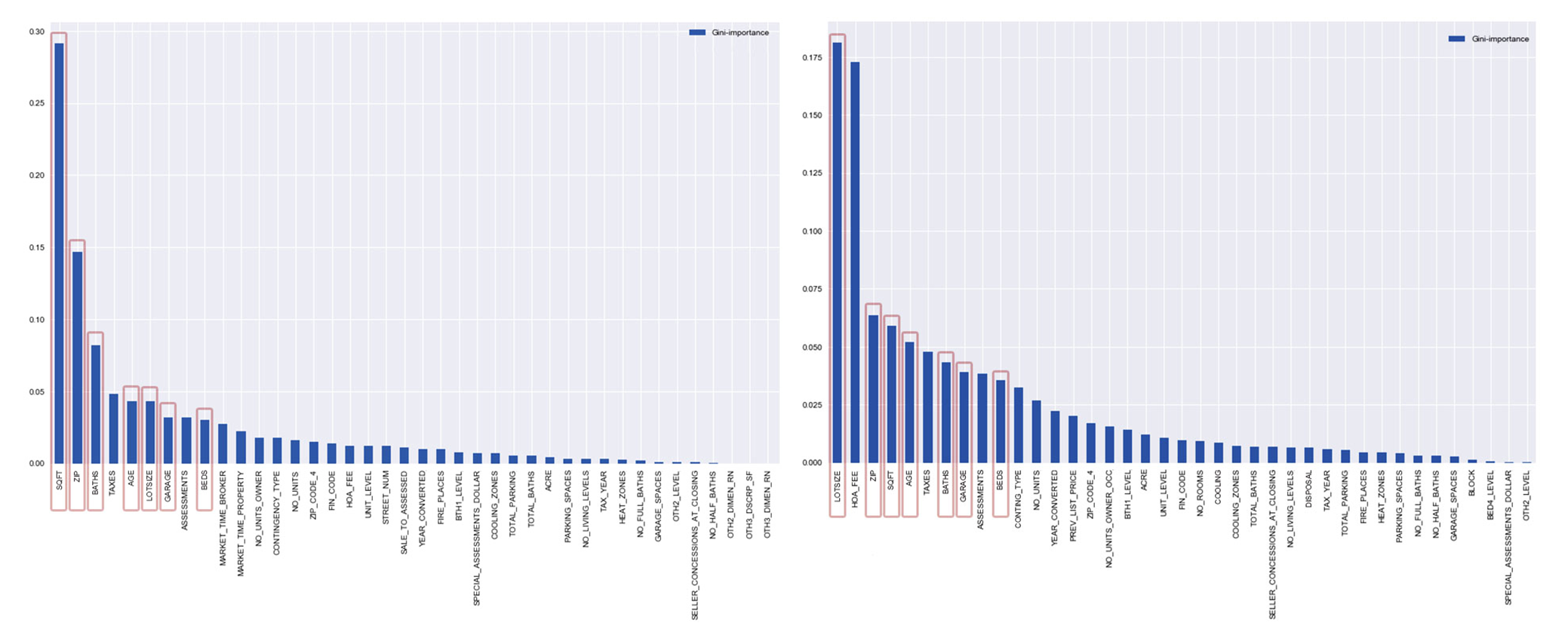}
    \caption{Feature importance graph for {\textit{price}} and {\textit{dom}}}
\end{figure*}

As expected, the location, square footage, and the number of bedrooms/bathrooms are among the strongest predictors. More surprising was the importance of binary features (e.g., garage or fireplace) and building-related fees. For the sake of simplicity of the model and the feature space, while also ensuring optimum performance at the same time, we decided to continue our analysis using the seven “basic” predictors: ZIP, LOTSIZE, AGE, BEDS, BATHS, GARAGE, and SQFT. We excluded one-hot encoded predictors due to the head-on capability of the CNN model (the extraction is described in subsection E). Furthermore, taxes and assessments, which were identified as important features, are established after a purchase takes place, making them false parameters in a given prediction scenario. 

The rest of the paper will focus on surpassing these seven "basic" predictors, (hereafter referred to as {\textit{MLS numeric features}}). The goal is to outperform the MLS numeric features with image features. 

\subsection{Image data}
Image data were gathered using three different  sources: MLS, Google Street View, and Google Maps. MLS images were split into interior and exterior images using the CNN described in subsection E. Exterior images were combined with Google Street View images. Finally, all images were regrouped and categorized as: 

\begin{itemize}
\item\textit{indoor}
\item\textit{outdoor} (exterior + Google Street View)
\item\textit{satellite} (Google Maps)
\end{itemize}

Together with the MLS numerical features, the total data corpus is represented in Table 2. Figure 2 gives an overview of all four image types, and provides visual clarification of the use of both outdoor and Street View images. 

\begin{figure}
    \includegraphics[width=\linewidth]{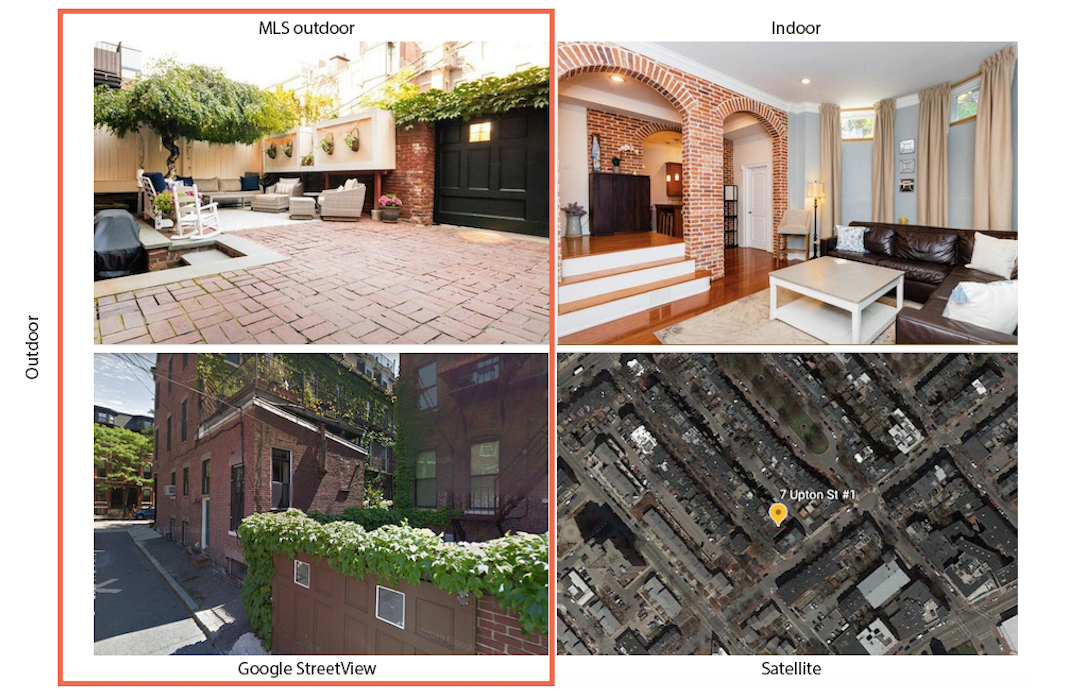}
    \caption{Different image types}
\end{figure}


\begingroup
\setlength{\tabcolsep}{1mm} 
\renewcommand{\arraystretch}{1} 
\begin{table}
    \scriptsize
    \begin{tabular}{p{4cm}p{4cm}}
    \makecell{
        \begin{tabular}{ c c }
            \hline
            MLS total units & 19,942 \\
            MLS features & 7 \\
            Response variables & 2 \\
            \hline
        \end{tabular}
        \\
        \\
        \begin{tabular}{p{2cm}p{1cm}}
            \hline
            Total images: & 399,120 \\
            Indoor: & 209,281 \\
            Outdoor: & 89,640 \\
            \hline
        \end{tabular}
        }
        &
        \begin{tabular}{p{1.4cm}p{2.2cm}}
            \hline
            MLS feature: & feature dsecription \\
            MLSNUM: & reference number \\
            PRICE: & final property price \\
            DOM: & days on the market \\
            ZIP: & zip code of the house \\
            BEDS: & number of bedrooms \\
            BATHS: & number of bathrooms \\
            LOTSIZE: & lot size of the house \\
            SQFT: & square footage \\
            GARAGE: & binary value \\
            AGE: & age of the property \\
            \hline
        \end{tabular}
    \\
    \\
    \multicolumn{2}{c}{
        \begin{tabular}{ c c c }
            \hline
            Indoor & [mls\_num\_img\_ind\_1, ..., mls\_num\_img\_ind\_10] & 10 \\
            Outdoor & [mls\_num\_img\_out\_1, ..., mls\_num\_img\_out\_10] & 10 \\
            Satellite & [mls\_num\_img\_sat\_15, ..., mls\_num\_img\_sat\_20] & 10 \\
            \hline
        \end{tabular}
        }
    \end{tabular}
    \\
    \caption{Metadata features with corresponding images per MLS listing}
\end{table}
\endgroup

\begin{figure*}
    \includegraphics[width=\textwidth]{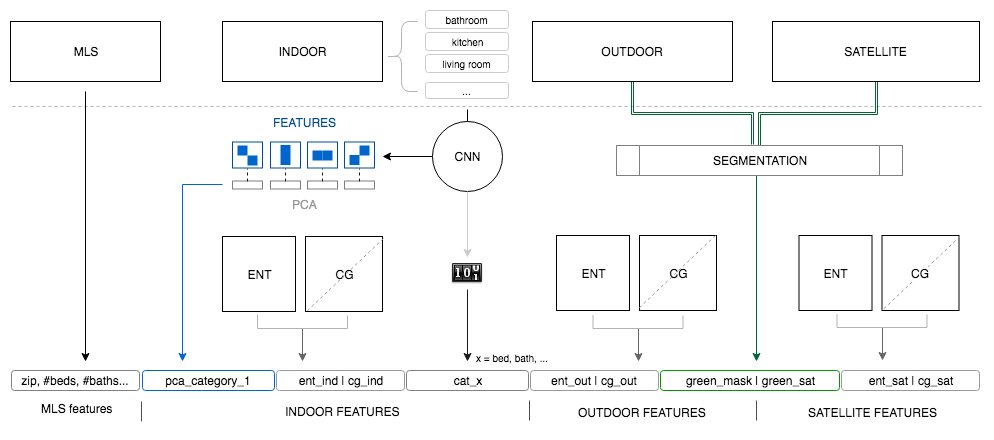}
    \caption{Feature extraction processes for indoor, outdoor, and satellite images. The architecture is using entropy, convolutional networks, and segmentation technique to abstract meaning into a single digit feature.}
\end{figure*}

The rest of the section focuses on individual image extraction techniques. Each subsection ends with the correlation coefficient estimates, plotting all features with respect to \textit{price} and \textit{dom}. Despite the fact that correlation does not necessarily imply causation, we wanted to compare the extracted predictors with the MLS numeric features before taking the final decision on acceptance or rejection. The corresponding correlation graphs for each subsection are given in the Appendix.

Figure 3 shows the process of extracting features using the different techniques described in this paper. The first part corresponds to the MLS data and the basic set of MLS features used. Then, the corresponding indoor, outdoor and satellite images (per unit) were processed and the features were extracted. The individual techniques and modeling details are presented in the following subsections.

\subsection{Entropy and Center of Gravity (CG)}
Information theory and Shannon's entropy \cite{Shannon1948} is the first method employed for extracting image features. By calculating the entropy of an image, we quantify its information potential. Particularly, we used the entropy to determine the quality \cite{SaliencyAttention} of the estate photos supplied, as well as the urbanization level of the estate's surrounding area, by using the satellite images. In the same manner, we used the center of gravity to quantify the quality (organization) of the photos.

Entropy is defined as an average quantity or a measure of the uncertainty of information. For a known probability, the entropy of an event is calculated by:

\begin{equation}
H=-\sum_{i=1}^{l}p_i\log_bp_i
\end{equation}

where p\textsubscript{i} represents the probability of occurrence of the symbol (the number of occurrences of the symbol divided by the number of symbols). Originally, Shannon's entropy was used to calculate the capacity of a communication channel, i.e. the amount of information that could be transferred. In this paper, we draw an analogy between the communication channel and the visual transfer of the information to the human observer \cite{TsaiLM08, Attneave54someinformational}. 

Another reason behind the use of the entropy measure was to quantify the amount of man-made formations by looking at satellite images. It has been found that entropy levels are very different for natural and artificial creations, with natural having lower levels. This might be perceived as counter-intuitive \cite{DBLP:journals/jdi/TsaiLM08}, since artificial formations are expected to have lower entropy. However, on the zoomed-out satellite images, the entropy of the natural content decreases, while the edges of the artificial content become closer, forcing the entropy to increase.

\begin{figure}[H]
    \includegraphics[width=250pt]{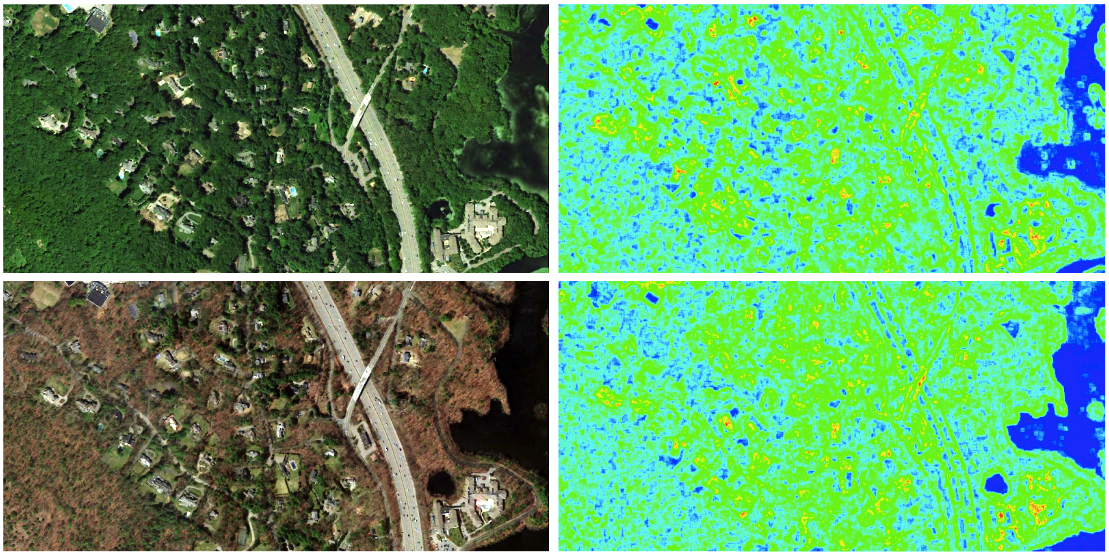}
    \caption{The season in which the images were taken and the level of greenness does not affect the entropy measure - the top (left) photo was taken during the summer, while the bottom (left) photo was taken during the fall. The entropy levels on right photos are almost identical.}
\end{figure}

Moreover, the entropy measure can be used for image segmentation purposes. For example, when measuring the level of greenness, the color saturation may vary between seasons. The satellite data on Google Maps and Google Street View are typically one to three years old. Given the variability and seasoning in such data, the potential of the entropy was very large, arising from the resistance color variations. Figure 4 shows that the entropy level (right) for both images is almost identical, despite the different seasons in which the satellite images were taken (summer in the upper image, fall in the lower).

\begin{figure}[H]
    \includegraphics[width=\linewidth]{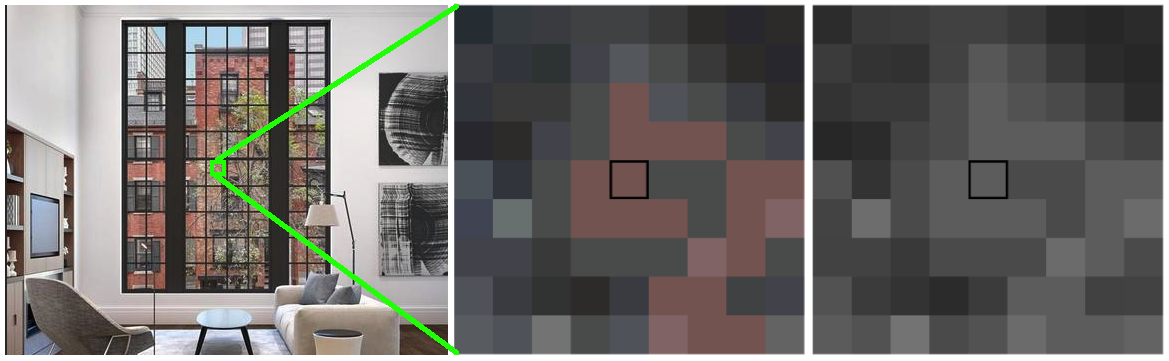}
    \caption{The image is converted into grayscale and each 9x9 pixel region is analyzed.}
\end{figure}

As expected, the value of local entropy varies based on the chosen window size. In general, the values obtained are represented by means of a binary logarithm, where one bit represents the quantity of information. This method allows for the use of a varying number of neighboring pixels, in a rather similar way to Shannon's approximation model. We calculated the entropy for each pixel against its surrounding 9x9 pixel matrix (Figure 5). The image is first converted to grayscale, and the probability of each grayscale intensity is then extracted. The lower the probability, the higher the entropy.

\begin{figure}
    \centering
    \includegraphics[width=\linewidth]{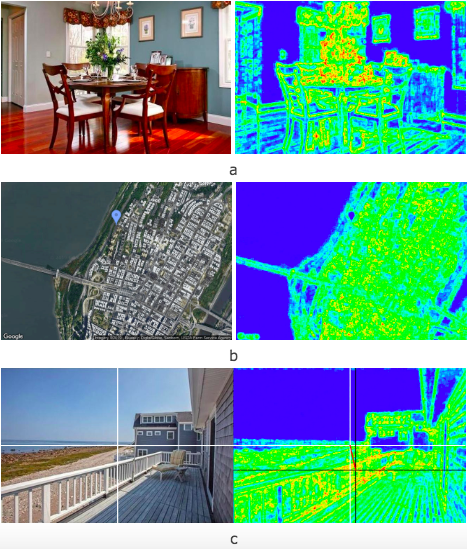}
    \caption{Entropy calculated for (a) the indoor image and (b) the satellite image; (c) the entropy with CG is calculated for the outdoor image.}
\end{figure}

The entropy measure was calculated for all indoor, outdoor, and satellite images. For the indoor and outdoor images, the entropy was used to determine the potential amount of information communicated, while for the satellite images, the entropy was used to calculate the level of urbanization for the area where the real estate is located. A visual representation of the calculated entropy for (a) indoor, (c) outdoor, and (b) satellite images (b) is given in Figure 6.

Additionally, for each image we calculated the distance between the center of the image and center of gravity (CG) (Figure 7 c). This measure is similar to the weighted centroid function, which calculates the average position of the grayscale intensity for an image. Instead of grayscale intensity, we used the pixels' entropy level. The horizontal coordinate of the center of gravity is calculated by the following formula:

\begin{equation}
\frac{\sum_{x=1}^{w}e_{x}\cdot {x}}{\sum_{x=1}^{w}e_{x}}
\end{equation}

where \textit{w} represents the image width, \textit{x} represents a column of pixels, and \textit{e\_x} represents the summarized entropy for that column of pixels. Accordingly, the vertical coordinate of the center of gravity is calculated by the following formula:

\begin{equation}
\frac{\sum_{y=1}^{h}e_{y}\cdot {y}}{\sum_{y=1}^{h}e_{y}}
\end{equation}

where \textit{h} represents the image height, \textit{y} represents a row of pixels, and \textit{e\_y} represents the summarized entropy for that row of pixels.

The distance between the CG and the center of the image is a scalar approximating the spread of information. The maximum value for this measure is the distance between the central pixel (width/2, height/2) and the corner pixel (0, 0).

We calculated two different entropy measures: the average entropy for the whole image (Figure 7 d) and the average entropy for nine regions (3x3 image matrices; Figure 7 c). Especially for the satellite images, average measures per segments were used to see if there was a correlation between the position (north/east/south/west) and the response variables. The idea behind the division into 3x3 regions was represent the property and its’ closest surroundings in the central region (approximately 500x500 ft), and to compare this against regions of equivalent size in the north, north-east, east, south-east, south, south-west, west, and north-west. 

\begin{figure}
    \includegraphics[width=\linewidth]{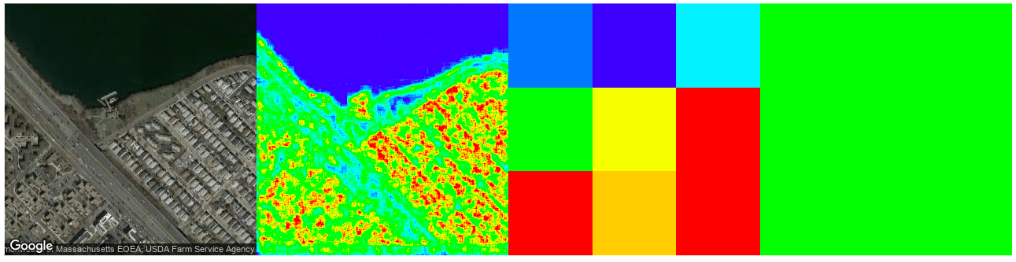}
    \caption{(a) Satellite image; (b) visualized entropy for the image; (c) average values for 3x3 regions; and (d) average value for an entire image.}
\end{figure}

The correlations between all features and response variables were also analyzed. Positive correlations were seen between \textit{price} and both the entropy extracted from the satellite images (average and per region, all zoom levels) and the CG. There is a significant negative correlation between the entropy and \textit{dom}, and the same trend was noticed for the indoor and outdoor images. Taking into account the observed trends, we decided to proceed with all extracted features from this subsection.

\subsection{Image Segmentation}
Our task in this part was to understand the size of the area surrounding the property. Expensive houses within a given city tend to exist in neighborhoods with large backyards and green spaces. For this specific task, we focused on outdoor and satellite images. Figure 8 gives an overview of the different image data with different levels of green areas surrounding the property. The distinction between the immediate surroundings for urban and rural properties is obvious.

Despite the fact that green areas are obvious to the human eye, extracting the percentage of green color could not be achieved by roughly calculating the frequency of green pixels (for example, the average proportions of green pixels in Figures 8a and 8b were approximately 30\% and 34\%, respectively). For successful image segmentation, we need to investigate the color space and see which option separates the color of interest. HSV (hue, saturation, value) has proven to be a good choice of color space for segmentation, usually with each axis representing one of the channels (for more details, see Figure 9).

\begin{figure}
    \begin{center}
    \includegraphics[width=225pt]{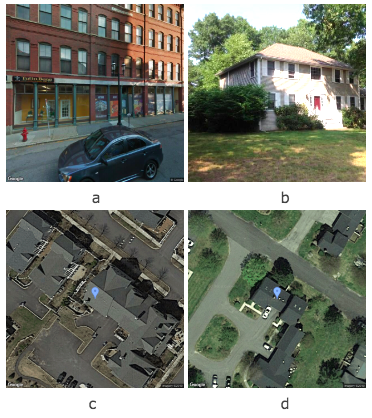}
    \caption{(a, b) Google Street View and (c, d) satellite images at zoom 20.}
    \end{center}
\end{figure}

Segmenting by color works well if channels are clearly separated, but not every saturation value separates the segments efficiently. In the process of discovering a range of good representations for outdoor vegetation, we produced color palettes based on the predominance of a color in an image (how much a particular color "dominates" an image). The range of extracted colors is given in Figure 9 (lower image).

This approach gives us a broad overview of the dominant colors in the entire dataset. Obviously, predominantly green images are those with the highest "greenness" scores. By using the most common representatives, we can generate different ranges and good masks for the entire dataset. To establish a set of colors that stand out in an image from a perceptual perspective, we used the k-means clustering algorithm \cite{Wu:2012:AKC:2344103}. K-means categorizes a set of data points into 'k' groups working on effective distance calculations in an unsupervised manner.

\begin{figure}
    \includegraphics[width=\linewidth]{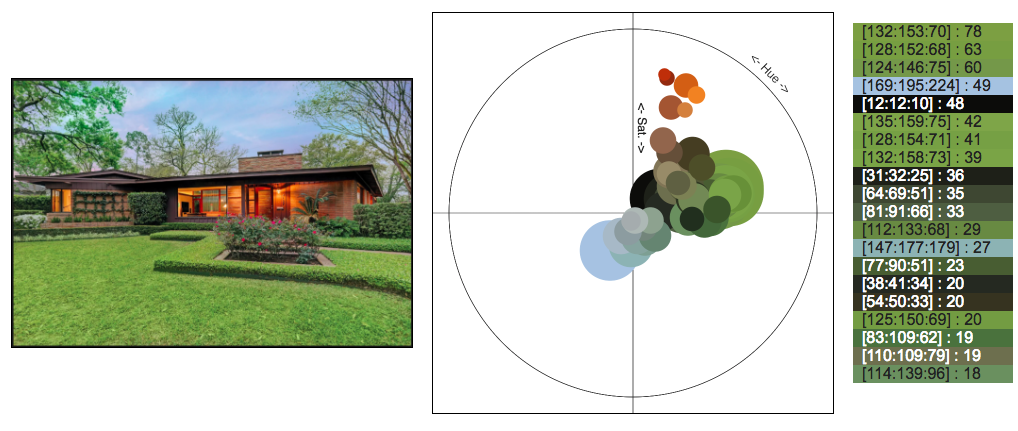}
    \caption{Using the technique in \cite{ColorProportionsWeb} to visualize dominant colors.}
\end{figure}

Once green color ranges (green masks) were established, they were applied as a threshold to the vegetation. The results of the color space segmentation technique are presented in Figure 10. On the left-hand side, images opened using the HSV color space are presented, and on the right-hand side are images with imposed masks. Using multiple masks, we can represent many different variations of green values captured in photos.

For each image in the outdoor and satellite datasets, we computed a "visual" area, populated by extracting segments as objects and estimating the percentage of total green space occupied. Finally, we averaged all "greenness" percentages per listing into a single \texttt{green\_mask} feature. The green mask \texttt{green\_sat} for satellite images proves to be the most efficient at a zoom level of 20.

After plotting the correlation coefficients, the \texttt{green\_mask} feature shows a strong negative correlation with \textit{dom}, capturing the "more green sells faster" trend. There is a strong impact of both features (\texttt{green\_mask} and \texttt{green\_sat}) on price and dom, as shown in the plots. We use these two features and report on their predictive power in the Evaluation section.

\begin{figure}
    \includegraphics[width=\linewidth]{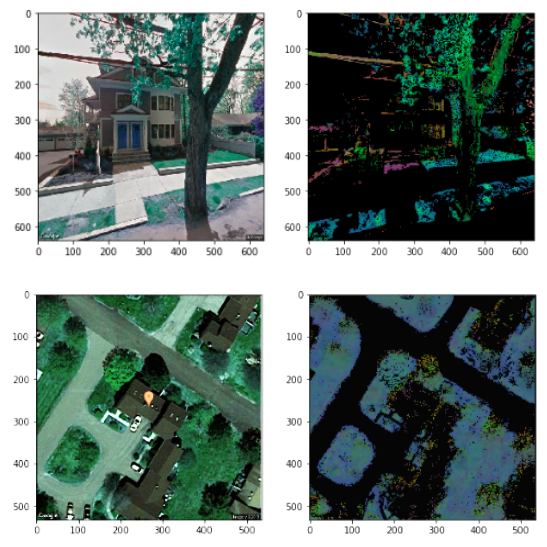}
    \caption{The results of color space segmentation}
\end{figure}

\subsection{Deep visual features}

\begin{figure*}
  \includegraphics[width=\textwidth]{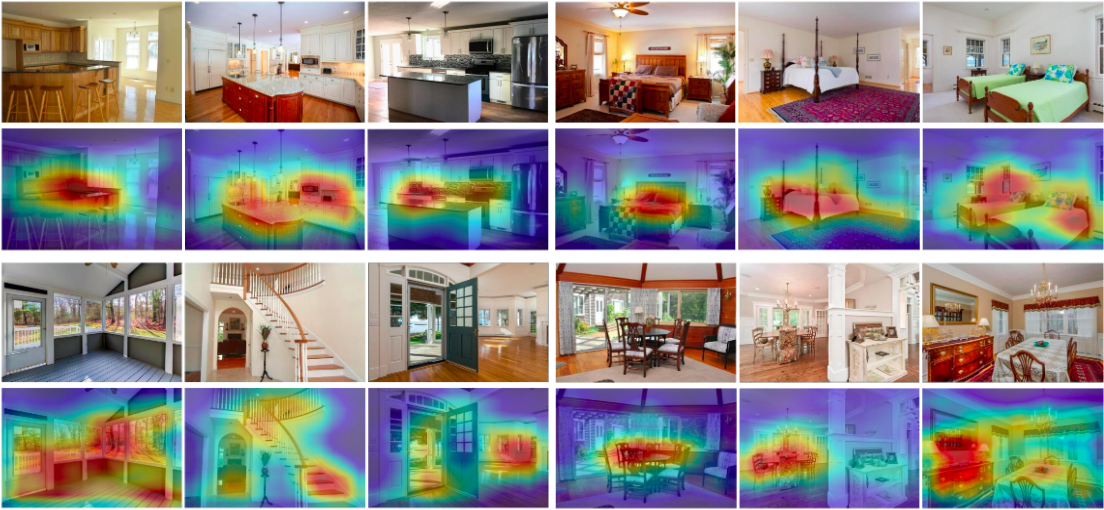}
  \caption{The activation maps show the different regions forming the focus of the CNN. The images have different styles that could account for "personal preferences" as well as \textit{price} and \textit{dom} ranges. We extract the category-related specifics that are the focus of the CNN and use them to model housing attractiveness. Extracting and comparing features gives style-based similarities between properties.}
\end{figure*}

For extracting deep image features, we first explored training the CNN network, both with and without employing weights, such as ImageNet \cite{5206848} and Places365 \cite{zhou2017places}. Later on, we explore the use of the pretrained architectures to extract image features directly. Both processes and corresponding results are presented in this section.  

After training different CNN architectures on the real-estate images, some results were obtained using ResNet50 \cite{DBLP:journals/corr/HeZRS15}. As mentioned previously, there is an inconsistent number of images per listing and thus, we had to focus only on categories presented with the most units. In order to organize the input data effectively, we created an ordered sequence of 9 images (kitchen, bathroom, bedroom, living room, dining room, satellite zoom 16, satellite zoom 18, outdoor 1, and outdoor 2). All images were size 1024 x 1024px, using the order presented on the figure 12. For the computational tasks in this research, we used an 8-core Intel Haswell 2.4GHz based system with 32GB of main memory, and 8TB of external memory.

\begin{figure}
    \includegraphics[width=\linewidth]{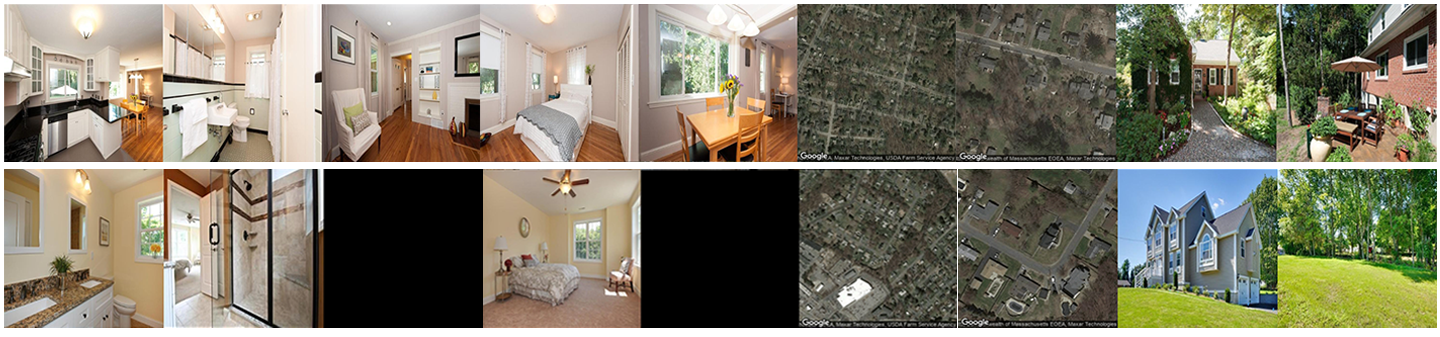}
    \caption{ Sequence of images with all categories present (top) and with a few categories missing (bottom)}
\end{figure}

ResNet50 was concatenated with the additional model into a fully-functional architecture, that takes care of variable data types. Results are presented with the Table 3. To emphasize the importance of previously extracted image features (Entropy, CG, and Image Segmentation), we incorporated these, alongside with the numeric mls features (shown as averages on the Table 3).


\begingroup
\setlength{\tabcolsep}{2pt} 
\renewcommand{\arraystretch}{1.5} 
\begin{table}
    \scriptsize
    \begin{tabular}{|p{4.35cm}|p{2cm}|p{2cm}|}
        \hline
        \textbf{Feature combinations} & \textbf{price (MAE)} & \textbf{dom (MAE)} \\
        \hline
        ResNet (images + metadata) & 0.28 & 0.79 \\
        \hline
        ResNet (images + metadata + averages) & 0.27 & 0.80 \\
        \hline
    \end{tabular}
    \\
    \caption{The results of the CNN trained on images and another numeric data}
\end{table}
\endgroup

The results clearly show poor performance of the CNN using the real-estate data. There are many reasons for it: One might be an insufficient data to train and test. Another could be hidden behind exploding gradients (neural networks for the regression tasks are especially prone to this problem). All these will be taken into account as the future venues of this research. 

More promising results were shown with pre-trained networks, used for classification problems. We processed all categories from the indoor images by the ResNet152-hybrid1365  \cite{DBLP:journals/corr/ZhouKLTO16} architecture, which is trained using a combination of ImageNet and Places365 datasets. This architecture is the most suitable pre-trained CNN, comprising a huge and diverse list of environments, with significant variety between classes for a large set of tasks. The ResNet152-hybrid1365 is based on the ResNet512 architecture, and predicts object and scene categories with 1,365 classes. It can be generalized and used on problems for which the network was not specifically trained. In the next subsection, we therefore explore a transfer learning technique that benefits from a pre-trained architecture, predefined weights and generalized features.

The deep image extraction pipeline is fairly straightforward. It consists of feature extraction followed by dimensionality reduction to narrow down a variable space. Just before the final layer, the CNN performs {\textit{global average pooling}}, which is used to discriminate between image regions presented as {\textit{class activation maps}} \cite{zhou2015cnnlocalization}. The activation maps for the kitchen, bedroom, entrance, and dining room categories are presented in Figure 11. The CNN focuses on the property-based specifics that are efficiently used to employ differences in styles between listings. 

Following the approach presented in \cite{Benefield2011}, we stored an additional figure, as a category representative (e.g., the number of bedroom images per listing). The number of images rarely accounts for a unit type, but it underlines the perceived quality of a specific section (more attractive property features tend to be presented multiple times). We extracted the most common set of categories observed in the majority of listings (such as the number of kitchen or living room images), excluding more unique and property-related features (e.g., wine cellar).

Preliminary analysis shows that the highest correlation is between the television room category and the \textit{price} of the listing. Moreover, certain features such as kitchen, dining room, and living room provide additional information in explaining both \textit{price} and \textit{dom}, probably on the grounds that no metadata describe them. We extracted a set of features (starting with \texttt{cat\_} as a category) as shown in Table 4.

Next, we proceed with deep extraction. We took features from the last CNN layer (before the network carries out the actual classification) and flattened them into a one-dimensional row. Unlike in \cite{7937942}, which used the average features of all the property-related images, we decided to (re)group them by category before the actual averaging took place. The main reason for this was to be able to identify a specific category-based style used to approximate the property interior attributes.

The extraction of deep features was followed by a Principal Component Analysis (PCA) dimensionality reduction \cite{doi:10.1098/rsta.2015.0202}. A total of 200 components were extracted per image, accounting for 85\% of the variability. We then averaged 200 PCA components for each individual category per unit listing. For example, one unit might have 15 images, with three images corresponding to the kitchen. For this specific scenario we would end up with [3, 200] PCA components that were finally flattened into [1, 200]. The components were called \emph{pca\_category}, where \emph{category} stands for \emph{bedroom}, \emph{bathroom}, \emph{kitchen} etc. The diagram in Figure 13 explains the entire process of (re)grouping, deep feature extraction, dimensionality reduction, and PCA averaging. 

\begin{figure}
    \includegraphics[width=\linewidth]{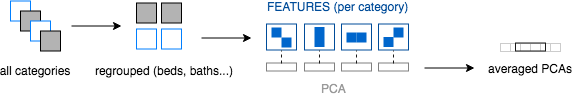}
    \caption{(Re)grouping, deep feature extraction, dimensionality reduction, and PCA averaging}
\end{figure}

As expected, plotting the first 15 components gives the highest correlations. The deep features seem to be effective in describing the interior style, showing significant correlations with \textit{dom}. From the graphs for \texttt{cat\_} features we can see a positive correlation with \textit{price} and a negative correlation with \textit{dom}. For the deep features, we decided to continue with the two components per category (the grouped bars on the graphs mainly correspond to the components taken from the same category). 

The final set of all image features is presented in Table 4. These will be used for performing the evaluation described in the next section.


\begingroup
\setlength{\tabcolsep}{2pt} 
\renewcommand{\arraystretch}{1.5} 
\begin{table}
    \scriptsize
    \begin{tabular}{|p{2cm}|p{6.35cm}|}
        \hline
       \textbf{Features} & \textbf{Description} \\
       \hline
       MLS & The basic information about the property: bed, bath, zip, lotsize, age, sqft, garage \\
       \hline
       \multicolumn{2}{c}{\textbf{Indoor features}} \\
       \hline
       PCA\_category\_x & 1st or 2nd component, x = \{ kitchen, bed, bath, living, basement, dinning \}  \\
       \hline
       ENT\_ind\_x & entropy measure, x = \{ t-top, m-middle, b-bottom, c-center, l-left, r-right \} \\
       \hline
       CG\_ind\_x & center of gravity, x = \{ distance, y-coordinate, x-coordinate \} \\
       \hline
       CAT\_x & x = \{ kitchen, bed, bath, living\_room, basement, dinning\_room \} \\
       \hline
       \multicolumn{2}{c}{\textbf{Outdoor features}} \\
       \hline
       GREEN\_mask & green mask for outdoor image \\
       \hline
       ENT\_out\_x & entropy measure, x = \{ t-top, m-middle, b-bottom, c-center, l-left, r-right \} \\
       \hline
       CG\_out\_x & center of gravity, x = \{ distance, y-coordinate, x-coordinate \} \\
       \hline
       \multicolumn{2}{c}{\textbf{Satellite features}} \\
       \hline
       GREEN\_sat & green mask for satellite image \\
       \hline
       ENT\_sat\_x & entropy measure, x = \{ t-top, m-middle, b-bottom, c-center, l-left, r-right \} \\
       \hline
       CG\_sat\_x & center of gravity, x = \{ distance, y-coordinate, x-coordinate \} \\
       \hline
    \end{tabular}
    \\
    \caption{ All image features extracted}
\end{table}
\endgroup

\section{Evaluation}
This section provides an explanatory analysis to evaluate the effectiveness of the different feature subsets and the models used. The response variables \textit{price} and \textit{dom}, both have extremely right-skewed distributions. A single log transformation allows the \textit{price} distribution to follow the normal distribution, while for the \textit{dom}, we bootstrapped the training dataset, and then performed a stratified sampling followed by a log transformation. Finally, \texttt{log\_price} and \texttt{log\_dom} were used as response variables. The evaluation metrics employed were the Mean Absolute Error (MAE) and Coefficient of Determination (R\textsuperscript{2}). Both of these are popular measures for evaluating the accuracy of boosting models, and are given by the equations:

\begin{equation}
MAE = \frac{1}{N}\sum_{i=1}^{N}\left | t_{i} - p_{i} \right |
\end{equation}

\begin{equation}
R^{2} = 1 - \frac{E\left [ (Y - y)^{2} \right ]}{E\left [ (y - \bar{y})^{2} \right ]}
\end{equation}

After extracting the features, we split the dataset into 70\% training and 30\% testing datasets. The initial idea of complementing MLS numeric features to "boost" the models with \textit{image features} was evaluated in this section. The base model therfore consisted solely of MLS numeric predictors. It was then gradually improved by adding different sets of image predictors. 

The hedonic model involves a regression of the house prices against the basic attributes of each house. Attributes that are hypothesized to contribute to the price of a house usually include land size, age, number of bedrooms, number of bathrooms and so on. These features correspond to our basic MLS features, and thus we proceed using Ordinary Least Square (OLS) regression as our baseline model. After running the OLS model, we noticed weak performance in comparison with boosted trees. Moreover, the model has no predictive power whatsoever for \textit{dom}. We tested a further approach using regularizer and ridge regression, mostly because this method is adaptive to high dimensions. More specifically, a ridge is capable of reducing the variability and improving the accuracy in the presence of multicollinearity (which is the problem we are dealing with, as described later in this section). However, ridge regression does not allow for variable selection, and it fails to provide a parsimonious model with few parameters. This model performed less accurately compared to boosted trees, which we use in the rest of this section.

Boosting algorithms are still more useful in the regime of limited training data, little training time, and little expertise in parameter tuning. The significance of these models has been seen in some of the most accurate real-estate prediction scenarios \cite{ZillowPrizeWebpage}. Thus, our determination of the predictive power of the extracted features started by introducing the entire feature space, using the three most significant boosting models: XGBoost (XGB)\cite{XGBoost}, LightGBM (LGB) \cite{NIPS2017_6907}, and CATBoost (CAT) \cite{CatBoost}. Interestingly, the LGB model performs better after adding all image features for both response variables, while XGB performs best on the small feature set, using MLS features only. CAT performs significantly less accurately in both cases (before or after adding the image features). For more details, see Table 5.  


\begingroup
\setlength{\tabcolsep}{2pt} 
\renewcommand{\arraystretch}{1.5} 
\begin{table}
    \scriptsize
    \begin{tabular}{|p{1.5cm}|p{1.5cm}|p{1.5cm}|p{1.5cm}|p{1.5cm}|}
        \hline
        & \multicolumn{2}{c|}{\textbf{MLS\_features (M})} & \multicolumn{2}{c|}{\textbf{M + IMG\_features}} \\
        \hline
        & price & dom & price & dom \\
        \hline
        \multicolumn{5}{c}{\textbf{XGB}} \\
        \hline
        MAE & 0.15 & 0.31 & 0.17 & 0.24 \\
        \hline
        ${R}^2$ & 0.85 & 0.59 & 0.83 & 0.75 \\
        \hline
        \multicolumn{5}{c}{\textbf{LGB}} \\
        \hline
        MAE & 0.16 & 0.28 & 0.16 & 0.23 \\
        \hline
        ${R}^2$ & 0.82 & 0.63 & 0.83 & 0.77 \\
        \hline
        \multicolumn{5}{c}{\textbf{CAT}} \\
        \hline
        MAE & 0.19 & 0.45 & 0.18 & 0.32 \\
        \hline
        ${R}^2$ & 0.66 & 0.03 & 0.68 & 0.51 \\
        \hline
    \end{tabular}
    \\
    \caption{Comparison of the base model vs all features added using different boosting models}
\end{table}
\endgroup

We assumed that the small accuracy improvement might be affected by a large feature space. Furthermore, it is always a good practice to remove any redundant features despite the fact that boosted trees are unaffected by multidimensionality or multicollinearity. We plotted multicollinearity matrices examining different feature sets, as presented with Figure 15. Following this, we performed feature selection analysis using all the proposed gradient boosting models. This method is used to perform subset selection, which improves the accuracy, while looking for good combinations in a high-dimensional feature space  \cite{Guyon:2003:IVF:944919.944968}. 

We used all the proposed boosting models to transform the training dataset into a subset with selected features. We then took a pre-trained model and used a threshold to decide which features to select. The set score seems to converge at about 30 features (the same set of features proved most efficient for predicting \textit{dom}). The best feature sets are: n=36 (XGB), n=40 (LGB), and n=10 (CAT). The order of the features was established after plotting the feature importance graph for each model, and taking the first \textit{n} predictors from a sorted list.

The next step determines the predictive power of the image features. We compare and present the results of our analysis using different subsets of predictors. We use MLS numeric features as our base model and continue to add new predictors reporting on the accuracy/error. From analyzing different subsets, we can identify some interesting combinations, as shown in Table 6.


\begingroup
\setlength{\tabcolsep}{2pt} 
\renewcommand{\arraystretch}{1.5} 
\begin{table}
    \scriptsize
    \begin{tabular}{|p{2.8cm}|p{0.8cm}|p{0.8cm}|p{0.8cm}|p{0.8cm}|p{0.8cm}|p{0.8cm}|}
        \hline
        \multirow{2}{*}{\textbf{FEATURE combinations}} & \multicolumn{2}{c|}{\textbf{best\_model}} & \multicolumn{2}{c|}{\textbf{price\_accuracy}} & \multicolumn{2}{c|}{\textbf{dom\_accuracy}} \\
        \cline{2-7}
        & \textbf{price} & \textbf{dom} & \textbf{MAE} & \textbf{${R}^2$} & \textbf{MAE} & \textbf{${R}^2$} \\
        \hline
        base\_1 & \multicolumn{2}{c|}{OLS} & 0.25 & 0.48 & 0.77 & 0.03 \\
        \hline
        base\_1 & \multicolumn{2}{c|}{RIDGE} & 0.24 & 0.77 & 0.76 & 0.04 \\
        \hline
        base\_1 & LGB & LGB & 0.16 & 0.821 & 0.29 & 0.62 \\
        \hline
        \textbf{base\_2} & XGB & LGB & \textbf{0.15} & \textbf{0.85} & \textbf{0.28} & \textbf{0.63} \\
        \hline
        base\_1 + outdoor features & LGB & LGB & 0.14 & 0.853 & 0.27 & 0.66 \\
        \hline
        base\_2 + indoor features & XGB & LGB & 0.14 & 0.854 & 0.27 & 0.68 \\
        \hline
        CAT best (n=10) & XGB & LGB & 0.217 & 0.68 & 0.26 & 0.7 \\
        \hline
        base\_2 + outdoor features & XGB & LGB & 0.14 & 0.854 & 0.26 & 0.7 \\
        \hline
        base\_2 + indoor and outdoor features & LGB & XGB & 0.13 & 0.862 & 0.26 & 0.7 \\
        \hline
        base\_2 + satellite features & LGB & LGB & 0.12 & 0.876 & \textbf{0.29} & \textbf{0.62} \\
        \hline
        XGB best (n=36) & LGB & LGB & 0.12 & 0.883 & 0.21 & 0.77 \\
        \hline
        \textbf{LGB} best (\textbf{n=40}) & LGB & LGB & \textbf{0.11} & \textbf{0.901} & \textbf{0.2} & \textbf{0.78} \\
        \hline
    \end{tabular}
    \\
    \caption{Accuracy for various models including MLS numeric and image features.}
\end{table}
\endgroup

Table 6 uses the following enumeration: \textit{base\_1} represents a base model using LOTSIZE, AGE, SQFT, ZIP, and BATHS as features. A model with the name \textit{base\_2} also adds BEDS and GARAGE into the feature space. These two combinations are complemented with image features. The XGB, LGB, and CAT combinations use all seven features from the \textit{base\_2} model. Some additional combinations can be observed, and all of these are presented in Table 6.

Furthermore, we tested our feature extraction technique using New York data, focusing on both different time-span as well as the region (table 7). Compared to the Boston data, the results are less satisfactory, however our approach still over-performs the baseline models. Assuming different trends, styles, and patterns with the New York data, we split the data into train/test and rerun the algorithms. Although, the results were significantly better (R-sq: 0.91, MAE: 0.1 for the \textit{price} and R-sq: 0.88, MAE: 0.12 for the \textit{dom}) we decided not to include them in the table 7, due to a small sample size. In all cases, LGB models performed the best. Satellite images alone do not bring any additional value, but improve the overall accuracy of the final model.


\begingroup
\setlength{\tabcolsep}{1mm} 
\renewcommand{\arraystretch}{1.5} 
\begin{table}
    \scriptsize
    \begin{tabular}{|p{2.22cm}|p{1.9cm}|p{1.9cm}|p{1.9cm}|}
        \hline
        & \textbf{\# listings} & \textbf{\# images} & \textbf{sold} \\
        \hline
        New York & 1,822 & 21,843 & jan-dec 2018 \\
        \hline
        \multicolumn{4}{|c|}{{\includegraphics[width=8cm]{photos/tbl7_fig1.png}}} \\
        \hline
    \end{tabular}
    \begin{tabular}{|p{2.9cm}|p{1.2cm}|p{1.2cm}|p{1.2cm}|p{1.2cm}|}
        \multirow{2}{*}{\textbf{FEATURE combinations}} & \multicolumn{2}{c|}{\textbf{price\_accuracy}} & \multicolumn{2}{c|}{\textbf{dom\_accuracy}} \\
        \cline{2-5}
        & \textbf{MAE} & \textbf{${R}^2$} & \textbf{MAE} & \textbf{${R}^2$} \\
        \hline
        base\_1 & 0.21 & 0.75 & 0.29 & 0.71 \\
        \hline
        base\_2 & 0.2 & 0.77 & 0.28 & 0.73 \\
        \hline
        base\_2 + indoor features & 0.15 & 0.854 & 0.19 & 0.79 \\
        \hline
        base\_2 + outdoor features & 0.15 & 0.854 & 0.19 & 0.79 \\
        \hline
        base\_2 + satellite features & \textit{0.22} & \textit{0.73} & \textit{0.29} & \textit{0.7} \\
        \hline
        base\_2 + indoor and outdoor features & 0.13 & 0.87 & 0.17 & 0.81 \\
        \hline
        XGB best (n=36 from Boston) & 0.12 & 0.88 & 0.17 & 0.82 \\
        \hline
        LGB best (n=40 from Boston) & 0.12 & 0.89 & 0.15 & 0.84 \\
        \hline
    \end{tabular}
    \\
    \caption{New York data description (top) and the evaluation (bottom).}
\end{table}
\endgroup

The role  the image features play in housing prediction is obvious, as these outperform basic predictors when modeling \textit{price} and \textit{dom}. As expected, the zip code is the most important feature for predicting both response variables. However, some of the previously set goals were achieved here: 

\begin{itemize}
    \item Some image features perform better than the MLS features 
    \item Image features contribute to the overall predictive power
\end{itemize}

The best feature combination includes the MLS numeric features combined with: \textit{green outdoor and satellite masks}, \textit{indoor and outdoor average entropy},  \textit{deep image features}, \textit{satellite regional entropy}, \textit{indoor and outdoor center of gravity}, and \textit{categorical features}. Some image features proved to be better predictors than the basic MLS features. Figure 14 provides more information regarding the individual importance of these features.

\begin{figure}
 \begin{center}
    \includegraphics[width=320pt, angle=90]{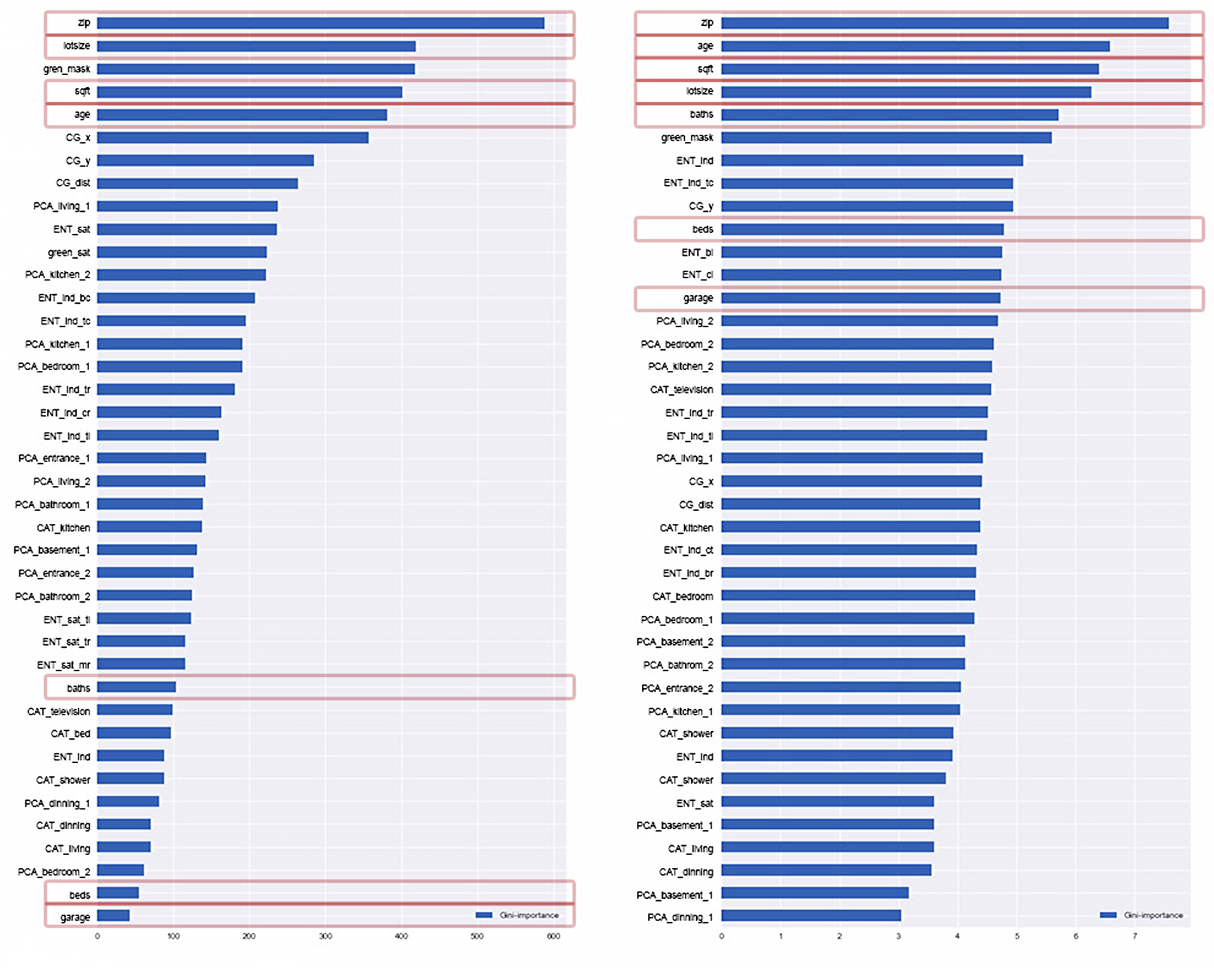}
     \end{center}
    \caption{Importance of all MLS numeric and image features.  MLS features are outlined.}
\end{figure}

We can observe that image features play a significant role in predicting \textit{price} and \textit{dom}. Furthermore, the LGB seems to be the best model, followed by the XGB. CAT did not perform well for any of the response variables. CAT is a different implementation of gradient boosting which at times can give slightly more accurate predictions, in particular when large amounts of categorical features are introduced. After including the categorical variables (in our case, only ZIP could be one-hot encoded), running CAT did not improve the score. Our dataset consists of all numerical values, demonstrating the low accuracy using CAT, even after performing feature selection. Boosting models such as XGB and LGB perform better even if they treat categorical variables as numeric values.

\begin{figure*}
    \includegraphics[width=\textwidth]{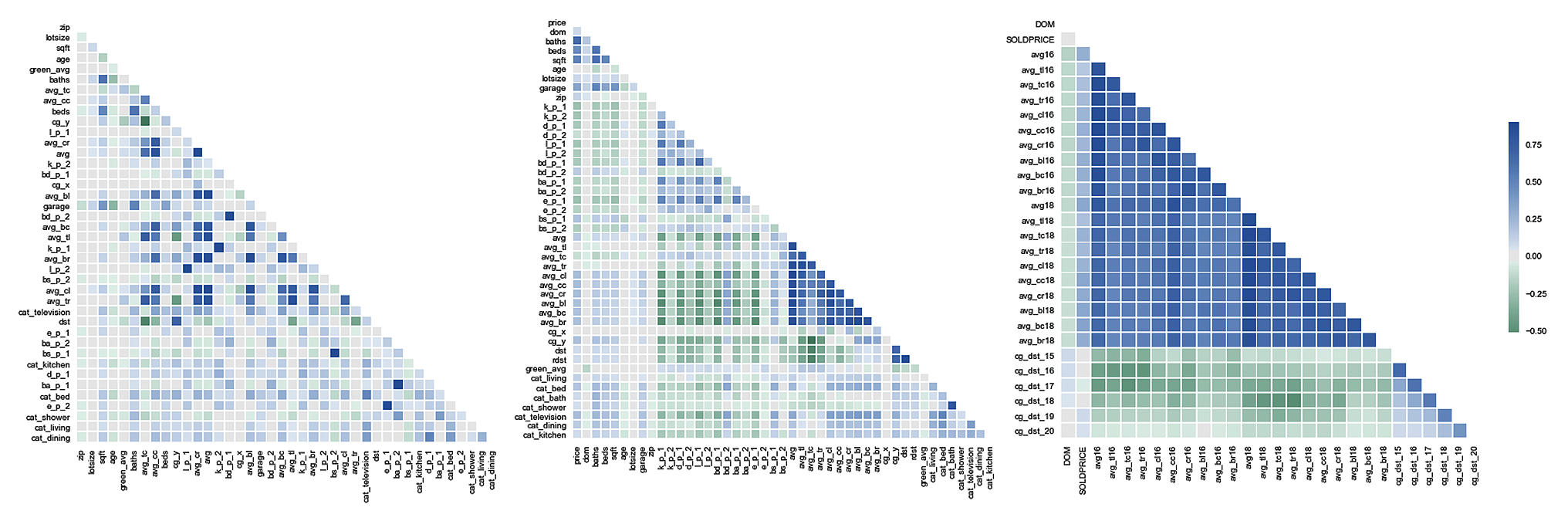}
    \caption{Groups of image features facing the multicollinearity problem. In the case of deriving new variables or dealing with large feature space, it is a good practice to remove any redundant features from the dataset followed by the importance graph, irrespective of the algorithm used.}
\end{figure*}

The best combination is a LGB model with n=40 features, which brings the R\textsuperscript{2} score to 90\% (MAE: 0.1). Predicting \textit{dom} proved efficient with the same model and set of features, showing a significantly lower score (MAE: 0.2, R\textsuperscript{2}: 0.78). This is probably due to a poor distribution (the dataset contains \textit{dom} values of between 0 and 1000 days on the market in a very skewed distribution). However, compared with the base model, LGB n=40 increases the value of R\textsuperscript{2} by 15\% for predicting \textit{dom}, and it is assumed that an even better score could be obtained in the case where the response variable follows a normal distribution. One interesting detail is that the satellite features alone do not contribute to the overall predictive power for estimating \textit{dom}. From the importance graph, it can be noticed that \textit{dom} is mainly explained by extracting information from the indoor images and based on the immediate surroundings captured in the outdoor images.

The feature importance graph is presented in Figure 14. Suggestion to the reader to bare in mind that the plot of \textit{price} (upper image) uses a power scale, due to the fact that the zip code drives the prediction power and other features are perceived as significantly less important. The importance graph for \textit{dom} (down) was not scaled. Both plots show the MLS numeric features enclosed with rectangles. From a comparison with Figure 1, the shift in the MLS numeric feature bars caused by the importance of image features can be seen.

Due to the nature of boosting models, we assume that further combinations into an ensemble would increase the overall accuracy. However, the examination of different model sets, the use of weights, and the introduction of different feature engineering techniques was outside the scope of this paper.

\section{Conclusion and Future Work}

In this paper, we prove the usefulness of image features in modeling real-estate attractiveness. A group of 40 features was selected to achieve successful content interpretation with the full set of techniques proposed. The research was based on all property-related images, using different visual perspectives, argue the that techniques proposed in this paper contribute to bringing subjectivity to boosting predictive algorithm.

The real estate appraisal process could benefit from the introduction of image processing into modeling, thus introducing a subjectivity factor into the evaluation process. Appraisers usually visit homes to evaluate both the interior and exterior of the properties. This process takes a lot of time, and creates uncertainty due to differences in human expertise. Due to a recently developed deep learning approach, models have become smart enough to interpret visual content in a way similar to human perception. Image features can easily be included as a quantification of an image and as a universal language, thus speeding up the appraisal process and making it more objective and fair.

Future avenues for the research presented in this paper will take different paths depending on the actual image type. First, we will decompose segments into a different image groups. Using various extraction techniques, we provide a set of combinations that can be easily re-combined and used in many similar scenarios. Next, the CNN is pre-trained on different places, and then needs to be re-trained on real estate data. Obtaining a larger set of images could make this possible. In case of a further employment of a transfer learning techniques, exploring different dimensionality reduction and clustering techniques might better differentiate or re-group listings based on similar criteria. 

The number of images per listing is not constant. We could employ different calculations for the weighted averages per listing. For outdoor images, we should take foliage into an account (different shades of brown and yellow) and/or perform a semantic segmentation. We find that it is important to distinguish between different types of green/yellow space to account for perceived attractiveness. Furthermore, the condition of a building could be better examined as an important driving factor in real estate modeling.

Machine learning methods can be used to analyze large datasets and conduct model selection in the context of causal inference. Issues such as multicollinearity across house attributes and incorrect functional forms threaten the underlying performance of hedonic models. Furthermore, a strong limitation of the majority of the models is a lack of consideration for endogeneity. Modern-day machine learning methods can accommodate both of these challenges \cite{JennyHo2016} and we plan to introduce some of the suggested techniques in future iterations.

Future work would also include deeper use of satellite images, an understanding of locality and neighborhood specifics as well as which types of green area (tree vs. grass) are captured with an image. Since the entropy results proved useful in segmenting large areas (e.g., water) in satellite images, we aim to estimate approximate distances in order to understand the location specifics of the surroundings of the property. Finally, we aim to train our CNN network to learn housing amenities from satellite images, and adapt it to automatically understand a location and neighboring characteristics.

\printbibliography

@electronic{ZillowPrizeWebpage,
 title     = "Zillow-Prize webpage",
 url       = "https://www.zillow.com/promo/zillow-prize/"
}

@electronic{ReportMLSPIN,
 title     = "Annual Report on the MLS PIN Housing Market",
 url       = "http://marketstatsreports.showingtime.com/MLSPIN_vnu43/MLSPIN_ANN_2018.pdf",
 year = 2018
}

@ARTICLE{1423983, 
author={Sung Young Jung and Jeong-Hee Hong and Taek-Soo Kim}, 
journal={IEEE Transactions on Knowledge and Data Engineering}, 
title={A statistical model for user preference}, 
year={2005}, 
volume={17}, 
number={6}, 
pages={834-843}, 
doi={10.1109/TKDE.2005.86}, 
ISSN={1041-4347}, 
month={June},}

@article{DBLP:journals/corr/HeZRS15,
  author    = {Kaiming He and
               Xiangyu Zhang and
               Shaoqing Ren and
               Jian Sun},
  title     = {Deep Residual Learning for Image Recognition},
  journal   = {CoRR},
  volume    = {abs/1512.03385},
  year      = {2015},
  url       = {http://arxiv.org/abs/1512.03385},
  archivePrefix = {arXiv},
  eprint    = {1512.03385},
  bibsource = {dblp computer science bibliography, https://dblp.org}
}

@ARTICLE{7937942, 
author={Q. {You} and R. {Pang} and L. {Cao} and J. {Luo}}, 
journal={IEEE Transactions on Multimedia}, 
title={Image-Based Appraisal of Real Estate Properties}, 
year={2017}, 
volume={19}, 
number={12}, 
pages={2751-2759}, 
doi={10.1109/TMM.2017.2710804}, 
ISSN={1520-9210}, 
month={Dec},}

@article{Wardrip11publictransit,
    author = {Keith Wardrip},
    title = {Public Transit's Impact on Housing Costs: A Review of the Literature},
    year = {2011}
}

@article{DBLP:journals/corr/PoursaeedMB17,
  author    = {Omid Poursaeed and
               Tomas Matera and
               Serge J. Belongie},
  title     = {Vision-based Real Estate Price Estimation},
  journal   = {CoRR},
  volume    = {abs/1707.05489},
  year      = {2017}
}

@Article{Dubin1999,
author="Dubin, Robin
and Pace, R. Kelley
and Thibodeau, Thomas G.",
title="Spatial Autoregression Techniques for Real Estate Data",
journal="Journal of Real Estate Literature",
year="1999",
month="Jan",
day="01",
volume="7",
number="1",
pages="79--96",
issn="1573-8809",
doi="10.1023/A:1008690521599",
url="https://doi.org/10.1023/A:1008690521599"
}

@article{Wang_2018,
	doi = {10.1088/1757-899x/322/5/052053},
	url = {https://doi.org/10.1088%2F1757-899x%2F322%2F5%2F052053},
	year = 2018,
	month = {mar},
	publisher = {{IOP} Publishing},
	volume = {322},
	pages = {052053},
	author = {Chengzhang Wang and Xiaoming Bai},
	title = {Boosting Learning Algorithm for Stock Price Forecasting},
	journal = {{IOP} Conference Series: Materials Science and Engineering}
}

@Inbook{Reichert2002,
author="Reichert, Alan K.",
editor="Wang, Ko
and Wolverton, Marvin L.",
title="Hedonic Modeling in Real Estate Appraisal: The Case of Environmental Damages Assessment",
bookTitle="Real Estate Valuation Theory",
year="2002",
publisher="Springer US",
address="Boston, MA",
pages="227--284",
isbn="978-1-4615-0909-7",
doi="10.1007/978-1-4615-0909-7_11",
url="https://doi.org/10.1007/978-1-4615-0909-7_11"
}

@InProceedings{10.1007/978-3-642-10677-4_58,
author="Nascimento, Diego S. C.
and Coelho, Andr{\'e} L. V.",
editor="Leung, Chi Sing
and Lee, Minho
and Chan, Jonathan H.",
title="Ensembling Heterogeneous Learning Models with Boosting",
booktitle="Neural Information Processing",
year="2009",
publisher="Springer Berlin Heidelberg",
address="Berlin, Heidelberg",
pages="512--519",
isbn="978-3-642-10677-4"
}

@article{StackingEnsemble,
  author    = {Sol Girouard and Zona Kostic},
  country   = {United States},
  title     = {Stacking Ensemble Approach for Combining Different Methods in Real Estate Prediction},
  volume    = {12},
  number    = {3},
  year      = {2018},
  pages     = {2767},
  ee        = {http://waset.org/abstracts/90597},
  url       = {http://waset.org/abstracts/Computer-and-Information-Engineering},
  bibsource = {http://waset.org/abstracts},
  conference = {ICCSDM 2018: International Conference on Computer Science and Data Mining, Miami, USA, (Mar 12-13,  2018)},
  issn      = {eISSN:1307-6892},
  publisher = {World Academy of Science, Engineering and Technology},
  index     = {International Science Index, Computer and Information Engineering,  12(3) 2018},
}

@article{Yoonseok2015,
  author = {Yoonseok Shin},
  title = {Application of Boosting Regression Trees to Preliminary Cost Estimation in Building Construction Projects},
  journal = {Computational Intelligence and Neuroscience},
  year = {2015},
  doi = {10.1155/2015/149702}
}

@article{DBLP:journals/corr/GuerinGTN17aa,
  author    = {Joris Gu{\'{e}}rin and
               Olivier Gibaru and
               St{\'{e}}phane Thiery and
               Eric Nyiri},
  title     = {{CNN} features are also great at unsupervised classification},
  journal   = {CoRR},
  volume    = {abs/1707.01700},
  year      = {2017}
}

@INPROCEEDINGS{7926625, 
author={A. J. {Bency} and S. {Rallapalli} and R. K. {Ganti} and M. {Srivatsa} and B. S. {Manjunath}}, 
booktitle={2017 IEEE Winter Conference on Applications of Computer Vision (WACV)}, 
title={Beyond Spatial Auto-Regressive Models: Predicting Housing Prices with Satellite Imagery}, 
year={2017}, 
volume={}, 
number={}, 
pages={320-329}, 
doi={10.1109/WACV.2017.42}, 
ISSN={}, 
month={March},}

@article{PredictionLiYu,
  title    = {Prediction on Housing Price Based on Deep Learning},
  author    = {Li Yu and  Chenlu Jiao and  Hongrun Xin and  Yan Wang and  Kaiyang Wang},
  country   = {China},
  institution={Renmin University of China},
    journal   = {International Journal of Computer, Electrical, Automation, Control and Information Engineering},  volume    = {12},
  number    = {2},
  year      = {2018},
  pages     = {90 - 99},
  ee        = {http://waset.org/publications/10008599},
  url       = {http://waset.org/Publications?p=134},
  bibsource = {http://waset.org/Publications},
  issn      = {eISSN:1307-6892},
  publisher = {World Academy of Science, Engineering and Technology},
  index     = {International Science Index 134, 2018},
}

@ARTICLE{2018arXiv180707155L,
       author = {{Law}, Stephen and {Paige}, Brooks and {Russell}, Chris},
        title = "{Take a Look Around: Using Street View and Satellite Images to Estimate House Prices}",
      journal = {arXiv e-prints},
         year = "2018",
        month = "Jul",
          eid = {arXiv:1807.07155},
        pages = {arXiv:1807.07155},
archivePrefix = {arXiv},
       eprint = {1807.07155},
 primaryClass = {econ.EM},
       adsurl = {https://ui.adsabs.harvard.edu/abs/2018arXiv180707155L}
}

@article{zhou2015cnnlocalization,
  title={{Learning Deep Features for Discriminative Localization.}},
  author={Zhou, B. and Khosla, A. and Lapedriza. A. and Oliva, A. and Torralba, A.},
  journal={CVPR},
  year={2016}
}

@article{SaliencyAttention,
    author = {Bruce, Neil D. B. and Tsotsos, John K.},
    title = "{Saliency, attention, and visual search: An information theoretic approach}",
    journal = {Journal of Vision},
    volume = {9},
    number = {3},
    pages = {5-5},
    year = {2009},
    month = {03},
    issn = {1534-7362},
    doi = {10.1167/9.3.5},
    url = {https://doi.org/10.1167/9.3.5},
    eprint = {https://jov.arvojournals.org/arvo/content\_public/journal/jov/933533/jov-9-3-5.pdf},
}

@ARTICLE{6875954, 
author={S. M. {Arietta} and A. A. {Efros} and R. {Ramamoorthi} and M. {Agrawala}}, 
journal={IEEE Transactions on Visualization and Computer Graphics}, 
title={City Forensics: Using Visual Elements to Predict Non-Visual City Attributes}, 
year={2014}, 
volume={20}, 
number={12}, 
pages={2624-2633}, 
doi={10.1109/TVCG.2014.2346446}, 
ISSN={1077-2626}, 
month={Dec},}

@INPROCEEDINGS{6909869, 
author={A. {Khosla} and B. {An} and J. J. {Lim} and A. {Torralba}}, 
booktitle={2014 IEEE Conference on Computer Vision and Pattern Recognition}, 
title={Looking Beyond the Visible Scene}, 
year={2014}, 
volume={}, 
number={}, 
pages={3710-3717}, 
doi={10.1109/CVPR.2014.474}, 
ISSN={1063-6919}, 
month={June},}

@Article{ijgi7030104,
AUTHOR = {Zhang, Yonglin and Dong, Rencai},
TITLE = {Impacts of Street-Visible Greenery on Housing Prices: Evidence from a Hedonic Price Model and a Massive Street View Image Dataset in Beijing},
JOURNAL = {ISPRS International Journal of Geo-Information},
VOLUME = {7},
YEAR = {2018},
NUMBER = {3},
ARTICLE-NUMBER = {104},
URL = {http://www.mdpi.com/2220-9964/7/3/104},
ISSN = {2220-9964},
DOI = {10.3390/ijgi7030104}
}

@article{RichardRYAIBlueBook,
  author = {Richard R. Yang and Steven Chen and Edward Chou},
  title = {AI Blue Book: Vehicle Price Prediction using Visual Features},
  journal = {CoRR},
  year = {2018}
}

@electronic{MLSSite,
 title     = "MLS Website",
 url       = "https://www.mlspin.com/"
}

@article{ViktoriyaISVFIUBF,
  author = {Viktoriya Krakovna and Jiong Du and Jun S. Liu},
  title = {Interpretable Selection and Visualization of Features and Interactions Using Bayesian Forests},
  journal = {CoRR},
  year = {2015}
}

@article{Shannon1948,
  author = {Shannon, Claude Elwood},
  doi = {10.1002/j.1538-7305.1948.tb01338.x},
  journal = {The Bell System Technical Journal},
  month = {7},
  number = 3,
  pages = {379--423},
  publisher = {Nokia Bell Labs},
  title = {A Mathematical Theory of Communication},
  url = {https://ieeexplore.ieee.org/document/6773024/},
  volume = 27,
  year = 1948
}

@article{TsaiLM08, author = {Du{-}Yih Tsai and Yongbum Lee and Eri Matsuyama}, title = {Information Entropy Measure for Evaluation of Image Quality}, journal = {J. Digital Imaging}, volume = {21}, number = {3}, pages = {338--347}, year = {2008}}

@electronic{ColorProportionsWeb,
 title     = "Color proportions of an image",
 url       = "http://www.geotests.net/couleurs/frequences_en.html"
}

@book{Wu:2012:AKC:2344103,
 author = {Wu, Junjie},
 title = {Advances in K-means Clustering: A Data Mining Thinking},
 year = {2012},
 isbn = {3642298060, 9783642298066},
 publisher = {Springer Publishing Company, Incorporated},
}

@INPROCEEDINGS{5206848, 
author={J. {Deng} and W. {Dong} and R. {Socher} and L. {Li} and and }, 
booktitle={2009 IEEE Conference on Computer Vision and Pattern Recognition}, 
title={ImageNet: A large-scale hierarchical image database}, 
year={2009}, 
volume={}, 
number={}, 
pages={248-255}, 
doi={10.1109/CVPR.2009.5206848}, 
ISSN={1063-6919}, 
month={June},}

@article{zhou2017places,
 title={Places: A 10 million Image Database for Scene Recognition},
 author={Zhou, Bolei and Lapedriza, Agata and Khosla, Aditya and Oliva, Aude and Torralba, Antonio},
 journal={IEEE Transactions on Pattern Analysis and Machine Intelligence},
 year={2017},
 publisher={IEEE}}

@Article{Benefield2011,
author="Benefield, Justin D.
and Cain, Christopher L.
and Johnson, Ken H.",
title="On the Relationship Between Property Price, Time-on-Market, and Photo Depictions in a Multiple Listing Service",
journal="The Journal of Real Estate Finance and Economics",
year="2011",
month="Oct",
day="01",
volume="43",
number="3",
pages="401--422",
issn="1573-045X",
doi="10.1007/s11146-009-9219-6",
url="https://doi.org/10.1007/s11146-009-9219-6"
}

@article{doi:10.1098/rsta.2015.0202,
author = {Ian T. Jolliffe  and Jorge Cadima },
title = {Principal component analysis: a review and recent developments},
journal = {Philosophical Transactions of the Royal Society A: Mathematical, Physical and Engineering Sciences},
volume = {374},
number = {2065},
pages = {20150202},
year = {2016},
doi = {10.1098/rsta.2015.0202}
}

@incollection{NIPS2017_6907,
title = {LightGBM: A Highly Efficient Gradient Boosting Decision Tree},
author = {Ke, Guolin and Meng, Qi and Finley, Thomas and Wang, Taifeng and Chen, Wei and Ma, Weidong and Ye, Qiwei and Liu, Tie-Yan},
booktitle = {Advances in Neural Information Processing Systems 30},
editor = {I. Guyon and U. V. Luxburg and S. Bengio and H. Wallach and R. Fergus and S. Vishwanathan and R. Garnett},
pages = {3146--3154},
year = {2017},
publisher = {Curran Associates, Inc.},
url = {http://papers.nips.cc/paper/6907-lightgbm-a-highly-efficient-gradient-boosting-decision-tree.pdf}
}

@article{CatBoost,
  author    = {Anna Veronika Dorogush and
               Vasily Ershov and
               Andrey Gulin},
  title     = {CatBoost: gradient boosting with categorical features support},
  journal   = {CoRR},
  volume    = {abs/1810.11363},
  year      = {2018}
}

@article{XGBoost,
  author    = {Tianqi Chen and
               Carlos Guestrin},
  title     = {XGBoost: {A} Scalable Tree Boosting System},
  journal   = {CoRR},
  volume    = {abs/1603.02754},
  year      = {2016}
}

@article{Guyon:2003:IVF:944919.944968,
 author = {Guyon, Isabelle and Elisseeff, Andr{\'e}},
 title = {An Introduction to Variable and Feature Selection},
 journal = {J. Mach. Learn. Res.},
 issue_date = {3/1/2003},
 volume = {3},
 month = mar,
 year = {2003},
 issn = {1532-4435},
 pages = {1157--1182},
 numpages = {26},
 url = {http://dl.acm.org/citation.cfm?id=944919.944968},
 acmid = {944968},
 publisher = {JMLR.org},
}

@article{Attneave54someinformational,
    author = {Fred Attneave},
    title = {Some informational aspects of visual perception},
    journal = {Psychol. Rev},
    year = {1954},
    pages = {183--193}
}

@article{DBLP:journals/jdi/TsaiLM08,
  author    = {Du{-}Yih Tsai and
               Yongbum Lee and
               Eri Matsuyama},
  title     = {Information Entropy Measure for Evaluation of Image Quality},
  journal   = {J. Digital Imaging},
  volume    = {21},
  number    = {3},
  pages     = {338--347},
  year      = {2008}
}

@article{doi:10.1177/03058298780070030601,
author = {Geoffrey Goodwin},
title ={Yet Another Paradigm?},
journal = {Millennium},
volume = {7},
number = {3},
pages = {251-259},
year = {1978},
doi = {10.1177/03058298780070030601},

URL = { 
        https://doi.org/10.1177/03058298780070030601
    
},
eprint = { 
        https://doi.org/10.1177/03058298780070030601
    
}

}

@article{JennyHo2016,
    author = {Jenny Ho},
    title = {Machine Learning for Causal Inference: An Application to Air Quality Impacts on House Prices},
    year = {2016}
}

@article{DBLP:journals/corr/ZhouKLTO16,
  author    = {Bolei Zhou and
               Aditya Khosla and
               {\`{A}}gata Lapedriza and
               Antonio Torralba and
               Aude Oliva},
  title     = {Places: An Image Database for Deep Scene Understanding},
  journal   = {CoRR},
  volume    = {abs/1610.02055},
  year      = {2016},
  url       = {http://arxiv.org/abs/1610.02055},
  archivePrefix = {arXiv},
  eprint    = {1610.02055},
  bibsource = {dblp computer science bibliography, https://dblp.org}
}

\begin{IEEEbiography}
    [{\includegraphics[width=1in,height=1.25in,clip,keepaspectratio]{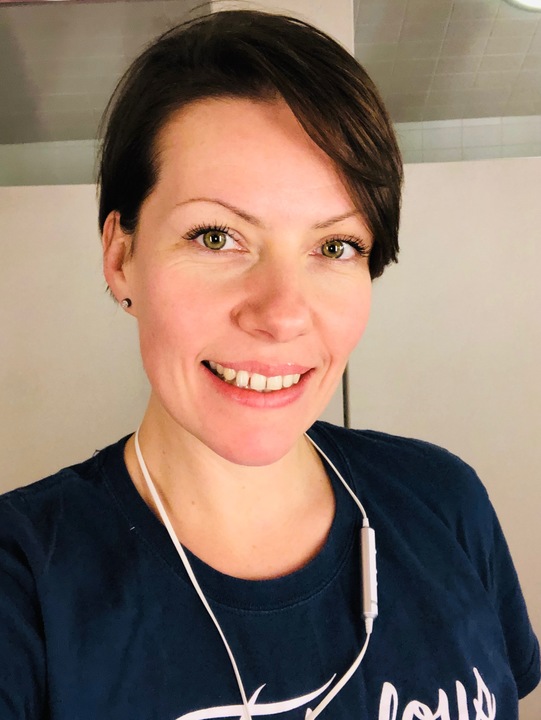}}]
    {Zona Kostic} is a research, teaching, and innovation fellow at Harvard University. Before coming to the US, Kostic hold an assistant professor position at the Faculty of Informatics and Computing in Belgrade, Serbia. In 2016, Kostic joined the Faculty of Arts and Sciences at Harvard University for a more empirical focus on data science, and in 2017, she obtained postdoc research position at the Visual Computing Group at Harvard SEAS, working on visual analytics projects. Kostic's recent advances focus on combining information visualization and machine learning into an intelligent Web systems, intensively collaborating with researchers inside and outside of the US. She has been a peer reviewer for numerous scientific journals as well as a committee member for most prestigious conferences. Kostic published six books and many research works at high impact journals.
\end{IEEEbiography}

\begin{IEEEbiography}
    [{\includegraphics[width=1in,height=1.25in,clip,keepaspectratio]{photos/ajevremovic.png}}]
    {Aleksandar Jevremovic} is a full professor at the Faculty of Informatics and Computing, Belgrade, Serbia, a guest lecturer at Harvard University in Cambridge, MA, and a visiting research fellow at the Cyprus Interaction Lab, Limassol, Cyprus. So far, he has authored/co-authored number of research papers and made contributions to three books about computer networks, computer network security and Web development. He is recognized as an Expert Level Instructor at Cisco Networking Academy program. Since 2018. he serves as a Serbian representative at the Technical Committee on Human\textemdash Computer Interaction of the UNESCO International Federation for Information Processing (IFIP).
\end{IEEEbiography}

\pagebreak
\raggedbottom

\appendices

\section*{Appendix}

\begin{center}

\begin{figure}[!h]
    \includegraphics[width=\linewidth]{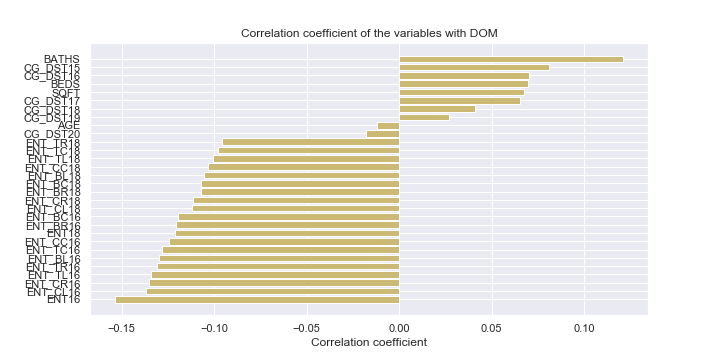}
    \caption{Correlation plot for entropy-based features extracted from satellite images and \textit{dom}}
\end{figure}

\begin{figure}[!h]
    \includegraphics[width=\linewidth]{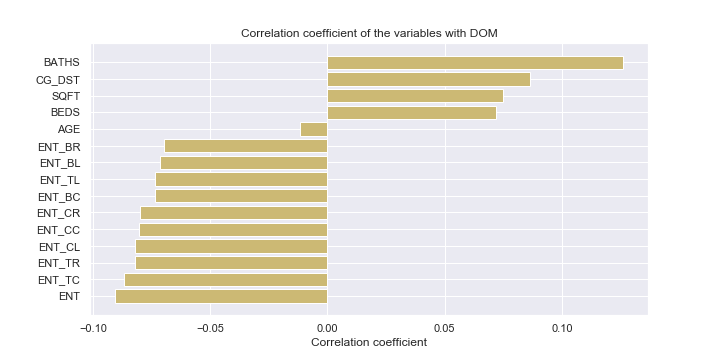}
    \caption{Correlation plot for entropy-based features extracted from outdoor images and \textit{dom}}
\end{figure}

\begin{figure}[!h]
    \includegraphics[width=\linewidth]{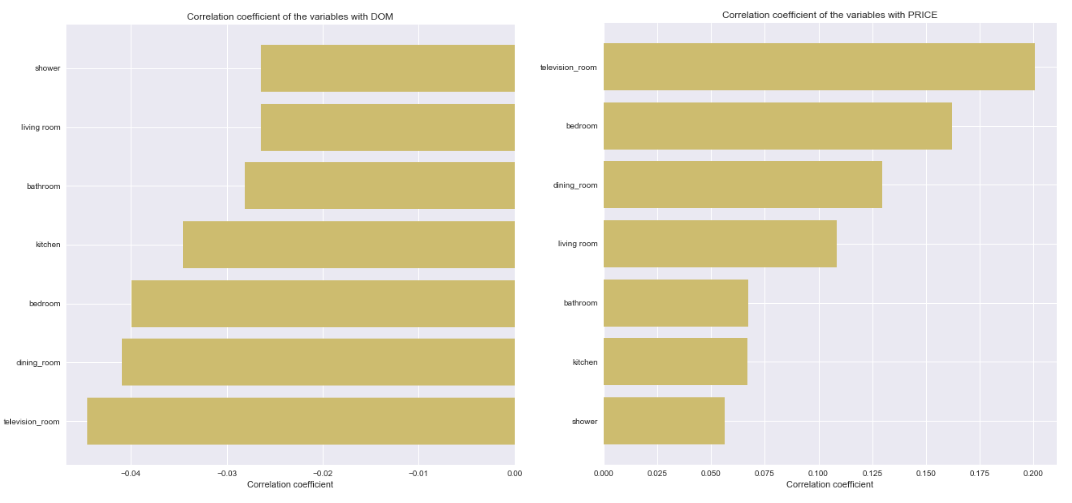}
    \caption{Correlation bars for \texttt{green\_avg} vs \texttt{green\_sat} features}
\end{figure}

\pagebreak
\raggedbottom

\begin{figure}[!h]
\vspace{10 mm}
    \includegraphics[width=\linewidth]{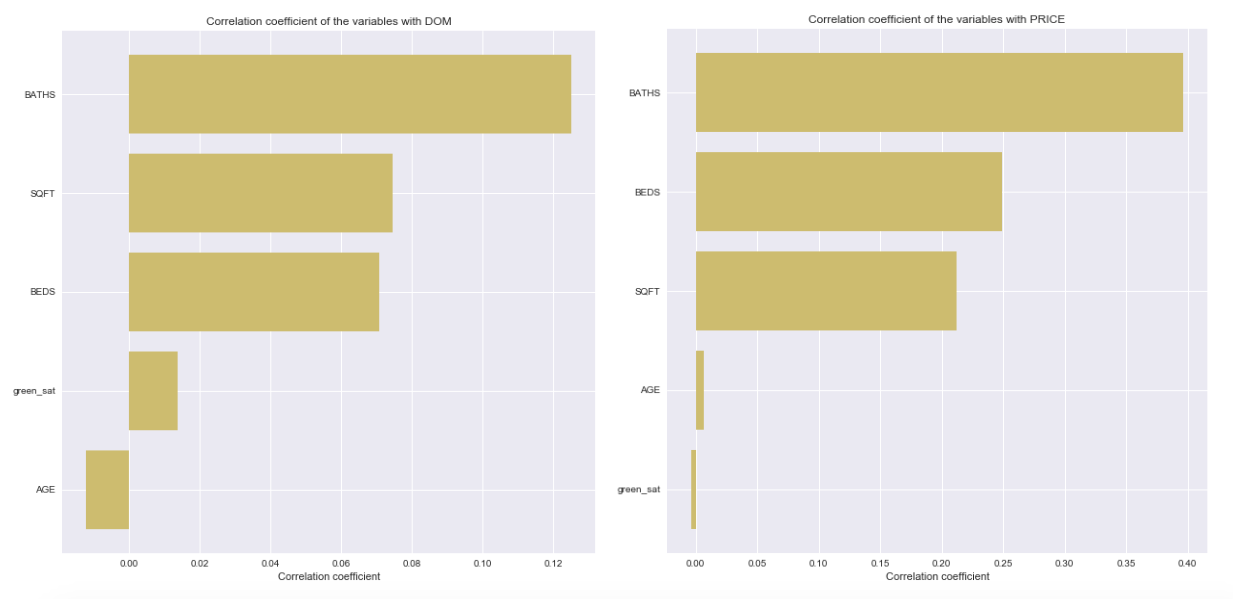}
    \caption{Correlation bars for \texttt{\_cat} features}
\end{figure}

\begin{figure}[!h]
\vspace{4 mm}
    \includegraphics[width=\linewidth]{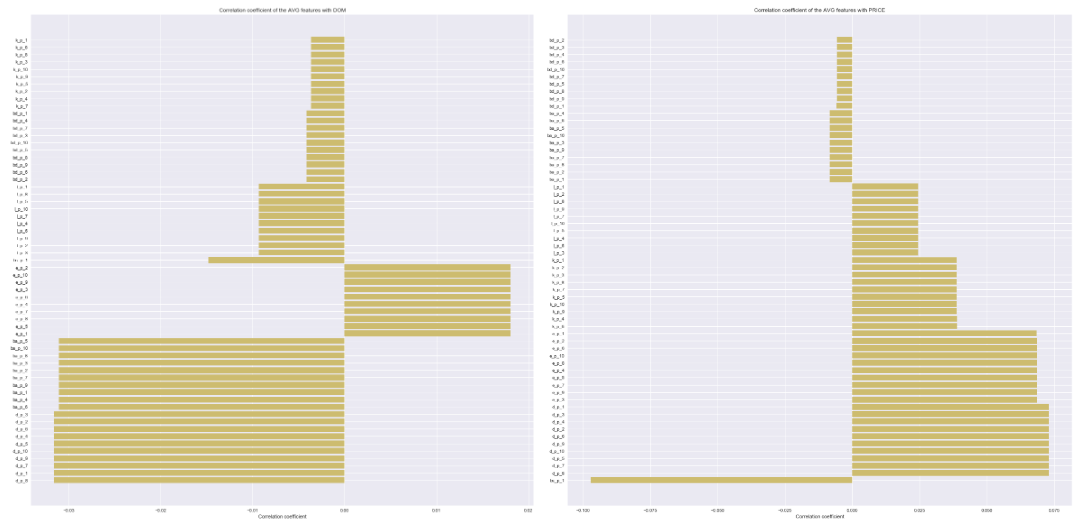}
    \caption{Correlation bars for first 15 deep image features (averaged PCA components)}
\end{figure}

\begin{figure}[!h]
    \includegraphics[width=\linewidth]{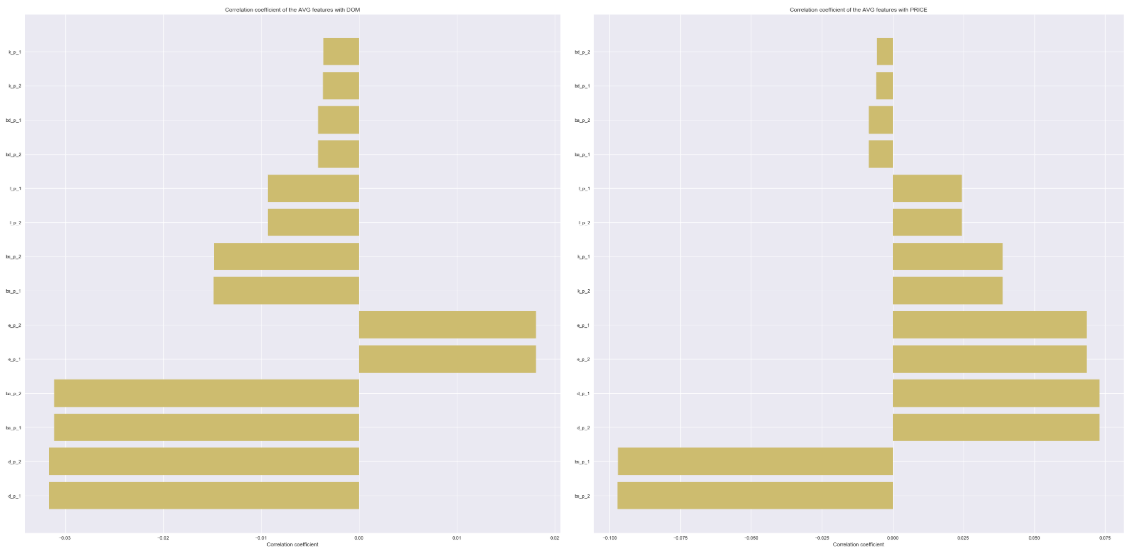}
    \caption{Correlation bars for first two deep image features (averaged PCA components)}
\end{figure}

\end{center}

\end{document}